\documentclass[11pt,a4paper]{article}
\usepackage[hyperref]{emnlp-ijcnlp-2019}
\usepackage{times}
\pdfoutput=1
\usepackage{latexsym}
\usepackage{url}

\aclfinalcopy % Uncomment this line for the final submission

\usepackage{xspace,mfirstuc,tabulary}

\usepackage{graphicx}
\usepackage{tabularx}
\usepackage{array}
\newcolumntype{P}[1]{>{\centering\arraybackslash}p{#1}}
\newcolumntype{Y}{>{\raggedright\arraybackslash}X}
\newcolumntype{?}{!{\vrule width 1pt}}
\newcolumntype{h}{!{\vrule width 0.7pt}}
\usepackage[fleqn]{amsmath}
\usepackage{amsfonts}
\usepackage{algorithm2e}
\usepackage[boldmath]{numprint}
\usepackage{url}
\usepackage[caption=false,font=footnotesize]{subfig}
\usepackage{float}
\usepackage{booktabs}
\usepackage{tikz}
\usepackage{todonotes}


\newcommand{\bs}{\boldsymbol} 
\newcommand{\Arrow}[1]{\parbox{#1}{\tikz{\draw[->](0,0)--(#1,0)}} }

\title{A Bayesian Approach for Sequence Tagging with Crowds}

\author{Edwin Simpson and Iryna Gurevych\\
  Ubiquitous Knowledge Processing Lab \\
  Department of Computer Science \\
  Technische Universit\"at Darmstadt \\
  \url{https://www.ukp.tu-darmstadt.de} \\
  {\tt \{simpson,gurevych\}@ukp.informatik.tu-darmstadt.de}
}

\begin{document}

\maketitle

% up to eight (8) pages of content, plus unlimited pages for references
% Each EMNLP 2018 submission can be accompanied by a single PDF appendix <--put equations in here
% + one .tgz or .zip archive containing software, and one .tgz or .zip archive containing data
% make sure that these are also anonymized.

% This paper shows how common annotator models for crowdsourcing can be seen as extensions of one another
% particularly in small-data scenarios that occur at the beginning of a crowdsourcing process.
%We publish our code to encourage adaptation and reuse.
\begin{abstract}
Current methods for sequence tagging, a core task in NLP, are data hungry,
which motivates the use of crowdsourcing as a cheap way to obtain labelled data.
However, annotators are often unreliable and current aggregation methods cannot capture common types of span annotation errors.
%, as they ignore sequential dependencies between annotators' labels.
To address this, we propose a Bayesian method for aggregating sequence tags
 that reduces errors by modelling sequential dependencies between the
 annotations as well as the ground-truth labels.
By taking a Bayesian approach, we account for uncertainty in the model due to both
annotator errors and 
the lack of data for modelling annotators who complete few tasks.
We evaluate our model on crowdsourced data for named entity recognition, information extraction and argument mining,
showing that our sequential model outperforms the previous state of the art. 
%and that 
%Bayesian approaches outperform non-Bayesian alternatives.
We also find that
 our approach can reduce crowdsourcing costs through
 more effective active learning, as it 
better captures uncertainty in the sequence labels when there are few annotations.
\end{abstract}

%%%%%%%%%%%%%%%%%%%%%%%%%%%%%%%%%%%%%%%%%%%%%%%%%%%%%%%%%%%%%%%%%

\section{Introduction}\label{sec:intro}

%The high demand for labelled training data in 
%A common NLP task that has benefited from deep learning is \emph{sequence tagging},
%which involves classifying sequences of tokens for tasks such as named entity recognition,
%part-of-speech tagging, or information extraction. 
Current methods for \emph{sequence tagging}, a core task in NLP,
use deep neural networks 
%sequence taggers 
that require
tens of thousands of labelled documents for training %containing hundreds of thousands of tokens
~\cite{ma2016end,lample2016neural}.
%This requirement for large labeled datasets 
This presents a challenge 
when facing new domains or tasks, where obtaining labels is often time-consuming or costly.
%Unlike I.I.D. classification tasks, such as document classification, the class labels in a sequence tagging task are dependent on the labels of the previous tokens.
Labelled data can be obtained cheaply by crowdsourcing, in which large
numbers of untrained workers annotate documents instead of more expensive experts.
For sequence tagging, this results in multiple sequences of unreliable labels for each document.
%Noisy labels could also be obtained from models trained on different domains,  
% multiple experts, or users of applications who click on and interact
%with text.

Probabilistic methods for aggregating crowdsourced data
have been shown to be more accurate than simple heuristics 
such as majority voting~\cite{Raykar2010,sheshadri2013square,rodrigues2013learning,hovy2013learning}.
However, existing methods for aggregating sequence labels
cannot model dependencies between the annotators' labels 
%miss error patterns such as a tendency to label overly long spans
~\cite{rodrigues2014sequence,nguyen2017aggregating}
and hence do not account for their effect on annotator noise and bias.
%%%% In this paper... we develop Bayesian model for aggregating sequence labels and compare different annotator models.
In this paper, we remedy this by proposing a sequential annotator model and applying it to
 tasks that follow a \emph{beginning, inside, outside} (\emph{BIO})
 scheme, 
in which the first token in a span of type `x' is labelled `B-x',  
subsequent tokens are labelled `I-x', 
and tokens outside spans are labelled `O'.
%This results in a sequence of labels, where each label depends on its predecessor.
%We propose an aggregation method that takes advantage of the sequential dependencies between BIO tags
%to learn the reliability of individual annotators and predict the true sequence.

When learning from noisy or small datasets, commonly-used
methods based on maximum likelihood estimation may produce over-confident predictions~\cite{xiong2011bayesian,srivastava2014dropout}. 
In contrast, Bayesian inference accounts for model uncertainty when making predictions.
%and enables hyperparameter tuning in unsupervised scenarios through Bayesian model selection~\cite{Bishop2006}. 
Unlike alternative methods that optimize the values for model parameters, Bayesian inference
integrates over all possible values of a parameter, weighted by a prior distribution that captures background knowledge.
The resulting posterior probabilities improve downstream decision making
as they include the probability of errors due to a lack of knowledge. For example, 
during active learning, posterior probabilities assist with selecting the most informative data points~\cite{settles2010active}.

In this paper, we develop \emph{Bayesian sequence combination (BSC)}, 
building on prior work that has demonstrated the advantages of Bayesian inference for aggregating unreliable classifications~\cite{kim2012bayesian,simpsonlong,Felt2016SemanticAA,paun2018comparing}.
BSC is 
the first fully-Bayesian method for aggregating sequence labels from multiple annotators.
As a core component of BSC, we also introduce the \emph{sequential confusion matrix (seq)},
a probabilistic model of annotator noise and bias, 
which goes beyond previous work by modelling sequential dependencies between annotators' labels.
Further contributions include a theoretical comparison of the probabilistic models of annotator noise and bias,
and an empirical evaluation on three sequence labelling tasks,
in which \emph{BSC} with \emph{seq} consistently outperforms the previous state of the art. 
We make 
%the modular implementation of our proposed method freely available, along with 
all of our code and data freely available\footnote{\url{http://github.com/ukplab/arxiv2018-bayesian-ensembles}}.

\section{Related Work}
%TODO: note that the HMM crowd study did not establish whether performance benefited from the text features,
% sequence model, or a combination. In our experiments we perform ablation and establish that the combination is needed to realise benefits of the sequential model.

%A number of works have investigated methods for aggregating non-sequential
%classifications from crowds.
Sheshadri and Lease~\shortcite{sheshadri2013square} benchmarked several 
aggregation models for non-sequential classifications, 
obtaining the most consistent performance from 
that of Raykar et al.~\shortcite{Raykar2010}, who
model the reliability of individual annotators
using probabilistic confusion matrices,
as proposed by Dawid and Skene~\shortcite{dawid_maximum_1979}.
%Nguyen et al.~\shortcite{hung2013evaluation} 
Simpson et al.~\shortcite{simpsonlong} showed that a Bayesian variant of
Dawid and Skene's model
can outperform maximum likelihood approaches and simple heuristics
when combining crowds of image annotators.
This Bayesian variant, independent Bayesian classifier combination
(\emph{IBCC})~\cite{kim2012bayesian}, 
was originally used to combine ensembles of automated classifiers
rather than human annotators. While
traditional ensemble methods such as boosting focus on how to generate
new classifiers~\cite{dietterich2000ensemble},
IBCC is concerned with modelling the reliability of each classifier in a given
set of classifiers.
To reduce the number of parameters in multi-class problems,
Hovy et al.~\shortcite{hovy2013learning} proposed \emph{MACE}, 
and showed that it performed better under a Bayesian treatment on NLP tasks.
Paun et al.~\shortcite{paun2018comparing} further illustrated
the advantages of Bayesian models of annotator ability on NLP classification tasks
with different levels of annotation sparsity and noise.

We expand this previous work by detailing the relationships between several annotator models
and extending them to sequential classification. 
% Could we mention this: The uncertainty in such inferences may then be
% used in applications such as jointly training clas-
% sifiers (Smyth et al., 1995; Raykar et al., 2010),
% comparing crowdsourcing systems (Lease and
% Kazai, 2011), or characterizing corpus accuracy
% (Passonneau and Carpenter, 2014).
% (OR maybe mention it in the intro paragraph).
% For classification tasks in NLP, 
%  evaluated six Bayesian methods,
% % pooled model -- assumes equal annotators but allows noise
% % DS, or rather my IBCC
% % MACE
% % hierarchical DS, basically original IBCC with pooled priors
% % item difficulty model, which ignores annotator reliability
% % logrndeff, item difficulty and annotator ability
% % Bayesian: uses Hamilonian Monte Carlo
% for their ability to obtain gold labels from crowdsourced data, model annotators and item difficulty.
% They found that...
% \cite{}. 
Here we focus on the core annotator representation, rather than extensions 
for clustering annotators~\cite{venanzi2014community,moreno_bayesian_2015},
modeling their dynamics~\cite{simpsonlong},
adapting to task difficulty~\cite{whitehill2009whose,bachrach2012grade},
or time spent by annotators~\cite{venanzi2016time}.

%Previous works have applied crowdsourcing to sequence labeling tasks such as named entity recognition~\cite{ritter2011named}
%and POS-tagging~\cite{hovy2014experiments}, but these do not develop new aggregation methods.
%To account for disagreement between annotators when training a sequence tagger, Plank et al. ~\shortcite{plank2014learning} 
%modify the loss function of the learner.
%to account for confusion between labels. 
%However, typical cross entropy loss
%naturally accommodates probabilities of labels as well as discrete labels~\cite{bekker2016training}.
%so it is unnecessary to adapt the loss function if  
%probabilities of training labels are used to account for disagreement between 
%multiple annotators
Methods for aggregating sequence labels 
include \emph{CRF-MA}~\cite{rodrigues2014sequence},
a CRF-based model that assumes only one annotator is correct for any given label.
Recently, Nguyen et al.~\shortcite{nguyen2017aggregating} proposed a hidden Markov model (HMM) approach that outperformed CRF-MA, called \emph{HMM-crowd}.
%HMM-crowd models the distribution of text tokens conditioned on the hidden state.
%More sophisticated LSTM-based sequence taggers must be trained separately given the true labels estimated by HMM-crowd.
Both CRF-MA and HMM-crowd use simpler annotator models than Dawid and Skene~\shortcite{dawid_maximum_1979}
that do not capture the effect of sequential dependencies on annotator reliability.
Neither CRF-MA nor HMM-crowd use a fully Bayesian approach,
which has been shown to be more effective for handling uncertainty
due to noise in crowdsourced data for non-sequential classification~\cite{kim2012bayesian,simpsonlong,venanzi2014community,moreno_bayesian_2015}.
In this paper, we develop a sequential annotator model and an approximate Bayesian method for aggregating sequence labels.  %that improves performance over previous approaches.

\section{ Modeling Sequential Annotators }\label{sec:annomodels}

When combining multiple annotators with varying skill levels, we can improve performance by modelling their individual noise and bias using a probabilistic
model.
Here, we describe several models 
that do not consider dependencies between annotations in a sequence,
before defining \emph{seq}, 
a new extension that captures sequential dependencies. 
Probabilistic annotator models 
each define a different function, $A$, 
for the likelihood that the annotator chooses label $c_{\tau}$
given the true label $t_{\tau}$, for the $\tau$th token in a sequence.

\textbf{Accuracy model (acc)}:
 the basis of several previous methods~\cite{donmez2010probabilistic,rodrigues2013learning},
\emph{acc} uses a single parameter for each 
annotator's accuracy, $\pi$: 
\begin{flalign}
 & A = p( c_{\tau} \! = \! i | t_{\tau} \! = \! j, \pi ) = \left.
\begin{cases}
  \pi  \!&\!\!\!\text{ where } i = j \\
  \frac{1 - \pi}{J-1} \!&\!\!\!\text{ otherwise}
\end{cases} 
\right\} \!, &&
\end{flalign}
where 
%$c_{\tau}$ is the label given by the annotator for token $\tau$, $t_{\tau}$ is its true label
%and 
$J$ is the number of classes.
%The limitation of this approach is that it
%It assumes reliability is constant,
%which means that 
This may be unsuitable when one class label dominates the data, 
since a spammer who always selects the most common label will nonetheless 
have a high $\pi$.
%despite their labels being uninformative.
%Annotator models define a likelihood... 
%\begin{flalign}
%& A = p(c_{\tau}, c_{\tau-1}, t_{\tau}) = p( c_{\tau} \!\!=\! i | c_{\tau-1}, t_{\tau} \!=\! j, \bs\pi ),&&
%\end{flalign}

\textbf{Spamming model (spam)}:
proposed as part of MACE~\cite{hovy2013learning}, this model also
assumes constant accuracy, $\pi$,
but that when an annotator is incorrect, they label according to 
a spamming distribution, $\bs\xi$, that is independent of the true label, $t_{\tau}$.
\begin{flalign}
A & = p( c_{\tau} = i | t_{\tau} = j, \pi, \bs\xi) && \nonumber \\
& = \left.
\begin{cases}
  \pi + (1 - \pi) \xi_j  &\text{ where } i = j \\
  (1 - \pi) \xi_j &\text{ otherwise}
\end{cases} 
\right\}.
\end{flalign}
This addresses the case where spammers choose the dominant label
% common label when the classes are imbalanced.
%While MACE can capture spamming patterns, 
but does not explicitly model 
different error rates in each class. 
%which
%may be an issue for sequence tagging using the 
%BIO encoding. For example, if an annotator frequently labels longer spans
% than the true spans by starting the spans early. In this 
% case, they may more frequently
%mis-label the `B' tokens than the `I' or `O' tokens,  which cannot be modelled by MACE. 
For example, if an annotator is better at detecting type `x' spans than type `y', or if they frequently miss the first token
in a span, thereby labelling the start of a span as `O' when the true label is `B-x', 
this would not be explicitly modelled by \emph{spam}.

\textbf{Confusion vector (CV)}: this approach learns a separate accuracy 
 for each class label~\cite{nguyen2017aggregating}
using parameter vector, $\bs\pi$, of size $J$:
\begin{flalign}
& A = p( c_{\tau} \!\!=\! i | t_{\tau} \!=\! j, \bs\pi ) = \left.
\begin{cases}
  \pi_j  \!\!\!\!\!\!&\text{ where } i \!=\! j \\
  \frac{1 \!- \!\pi_j}{J-1} \!\!\!\!\!\!&\text{ otherwise}
\end{cases} 
\! \right\} \!.&&
\end{flalign}
%For the incorrect label cases where $i \! \neq \! j$,
% $p( c_{\tau} \!\!=\! i | t_{\tau} \!=\! j, \bs\pi )$ is constant for all values of $i$.
% Therefore, t
This model does not capture spamming
patterns where one of the incorrect labels has a much higher likelihood than the others.

\textbf{Confusion matrix (CM)}~\cite{dawid_maximum_1979}:
this model can be seen as an expansion of the confusion vector so that $\bs\pi$ becomes a 
$J\times J$ matrix with values given by:
\begin{flalign}
& A = p( c_{\tau} \!\!=\! i | t_{\tau} \!=\! j, \bs\pi ) = 
  \pi_{j,i} .&&
\end{flalign}
This requires a larger number of parameters, $J^2$, compared to the $J+1$ parameters of MACE or $J$ parameters
of the confusion vector.
Like \emph{spam}, 
\emph{CM} %represents the probability of each mistake, so it 
can model spammers who frequently chose one label regardless
of the ground truth, but also
models different error rates and biases for each class.
%type of `B-x', `I-x' and `O' label.
However, \emph{CM} ignores dependencies between annotations in a sequence, % that affect these probabilities.
such as the fact that an `I' cannot immediately follow an `O'.
% Consider the following example where this may be a problem: three annotators produce sequences of labels as follows:
% O-B-I-I-I-O
% O-O-B-I-O-O
% O-O-O-O-O-O
% We can see that the first two annotators agree that the third token is part of the span, 

\textbf{Sequential Confusion Matrix (seq)}: we introduce a new extension to the confusion matrix to model the dependency 
of each label in a sequence on its predecessor,
giving the following likelihood:
\begin{flalign}
& A = p( c_{\tau} \!\!=\! i | c_{\tau-1} \!=\! \iota, t_{\tau} \!=\! j, \bs\pi ) = 
  \pi_{j,\iota,i} ,&&
\end{flalign}
where $\bs\pi$ is now three-dimensional with size $J\times J\times J$.
In the case of disallowed transitions, e.g. from $c_{\tau-1}=$`O' to $c_{\tau}=$`I', the value $\pi_{j,c_{\tau-1},c_{\tau}}\approx 0$, $\forall j$
is fixed \textit{a priori}. 
The sequential model can capture phenomena such as a tendency toward overly long sequences, by learning that I is more likely to follow another I, so that
$\pi_{O,I,I} > \pi_{O,I,O}$.
A tendency to split spans by inserting `B' in place of `I' can be modelled
by increasing the value of
$\pi_{I,I,B}$ without affecting $\pi_{I,B,B}$ and $\pi_{I,O,B}$.

The annotator models presented in this section 
include the most widespread models for NLP annotation tasks, 
and can be seen as extensions of one another.
The choice of annotator model for a particular annotator
 depends on the developer's understanding of the annotation task: 
 if the annotations have sequential dependencies, this suggests the \emph{seq} model;
for non-sequential classifications \emph{CM} may be effective with small ($\leq 5$) 
numbers of classes; \emph{spam} may be more suitable if there are many classes, as the number of parameters to learn is low. 
However, there is also a trade-off between the expressiveness of the model and the
number of parameters that must be learned. Simpler models with fewer parameters
may be effective if there are only small numbers of annotations from each annotator. 
%Our experiments in Section \ref{sec:expts_all} investigate this trade-off on NLP tasks involving sequential annotation.
The next section shows how these annotator models can be used as components of 
a complete model for aggregating sequential annotations. 
%The experiments in Section \ref{sec:expts_all} 
% test whether the more expressive seq annotator model,
%which has more parameters to learn, is beneficial in a realistic setting.

\section{A Generative Model for Bayesian Sequence Combination}\label{sec:model}
%TODO do we state the obvious: that the annotator models effectively weight good annotators more heavily and discard spammers?
%The generative story for our approach,
To construct a generative model for \emph{Bayesian sequence combination (BSC)}, 
we first define a hidden Markov model (HMM)
with states $t_{n,\tau}$ and observations $x_{n,\tau}$
using categorical distributions:
\begin{flalign}
t_{n,\tau} & \sim \mathrm{Cat}(\bs T_{t_{n,\tau-1}}), \\
x_{n,\tau} & \sim \mathrm{Cat}(\bs\rho_{t_{n,\tau}}), 
\end{flalign}
where $\bs T_j$ is a row of a transition matrix $\bs T$, and $\bs\rho_j$ 
is a vector of observation likelihoods for state $j$.
For text tagging, $n$ indicates a document and $\tau$ a token index, while
each state $t_{n,\tau}$ is a true sequence label
and $x_{n,\tau}$ is a token.
To provide a Bayesian treatment, we assume that 
$\bs T_j$ and $\bs\rho_j$ have Dirichlet distribution priors as follows:
\begin{flalign}
T_j & \sim \mathrm{Dir}(\bs \gamma_j), \hspace{1.0cm}
\bs\rho_j \sim \mathrm{Dir}(\bs \kappa_j), &
\end{flalign}
where $\bs \gamma_j$ and $\bs \kappa_j$ are hyperparameters.

Next, we assume one of the annotator 
models described in Section \ref{sec:annomodels} for each of $K$ annotators.
Selecting an annotator model is a design choice,
and all can be coupled with the Bayesian HMM above to form
a complete BSC model. In our
experiments in Section \ref{sec:expts_all}, 
we compare different choices of annotator model as components of BSC.
%We draw the parameters of the annotator models as follows:
%%The number of parameters depends on the choice of annotator model:
%%( t_{\tau}, c_{\tau}, c_{\tau-1})$.
%for \emph{acc}, only one parameter, $\pi^{(k)}$, is drawn per annotator $k$;
%for \emph{MACE}, we draw a single value $\pi^{(k)}$ and a vector $\xi^{(k)}$ of length $J$, 
%while for \emph{CV} we draw $J$ independent values of $\pi_j^{(k)}$, 
%and for \emph{CM}  
%we draw a vector $\bs\pi^{(k)}_j$ of size $J$ for each true label value $j\in \{1,...,J\}$; in the case of \emph{seq}, 
%we draw vectors $\bs\pi^{(k)}_{j,\iota}$ for each true label value 
%for each previous label value, $\iota$.\
All the parameters of these annotator models are probabilities,
so to provide a Bayesian treatment, we assume that they have Dirichlet priors. 
For annotator $k$'s annotator model, we refer to the hyperparameters
of its Dirichlet prior as $\bs\alpha^{(k)}$.
%As shown in Section \ref{sec:annomodels}, 
The annotator model defines a categorical likelihood
over each annotation, $\bs c^{(k)}_{n,\tau}$:
%$A^{(k)}(t_{n,\tau}, \bs c^{(k)}_{n,\tau}, \bs c^{(k)}_{n,\tau-1})$, where $\bs c^{(k)}_{n,\tau}$ is the $\tau$th label of document $n$.
%%The argument $\bs c_{n,\tau-1}$ is only required if $A^{(k)}$ is an instance
%%of \emph{seq} and is ignored by the other annotator models.
%We draw annotator $k$'s label $c^{(k)}_{n,\tau}$ 
%for each token $\tau$ in each document $n$ 
%according to a categorical distribution:
\begin{flalign}
& c^{(k)}_{n,\tau} \sim \mathrm{Cat}( [A^{(k)}(t_{n,\tau}, 1, \bs c_{n,\tau-1}^{(k)}), ..., & \nonumber \\
& \hspace{3cm} A^{(k)}(t_{n,\tau}, J, \bs c_{n,\tau-1}^{(k)}) ]). &
\end{flalign}

The annotators are assumed to be conditionally independent of one another given the true labels,
$\bs t$, which means that their errors are assumed to be uncorrelated. This is a strong assumption
when considering that the annotators have to make their decisions based
on the same input data. However, in practice, dependencies do not usually cause the 
most probable label to change~\citep{zhang2004optimality}, hence the performance of classifier combination methods 
is only slightly degraded, while avoiding the complexity of modelling dependencies between annotators~\citep{kim2012bayesian}.

\textbf{Joint distribution}: the complete model can be represented by the
joint distribution, given by:
\begin{flalign}
& p(\bs t, \bs A, \bs T, \bs\rho, \bs c, \bs x | \bs \alpha^{(1)},..., \bs \alpha^{(K)}, \bs\gamma,
\bs \kappa ) &  \\
%  & \approx q(\bs t, \bs A, \bs B, \bs\A^{(1)},...,\bs\A^{(K)},\bs\A^{(1)},...,\bs\A^{(S)}, \bs d^{(1)}, ...,\bs d^{(S)}) = q(\bs B) \prod_{n=1}^N q(\bs t_n) & \nonumber \\
& = \prod_{k=1}^K \left\{ p(A^{(k)} | \bs \alpha^{(k)}) \prod_{n=1}^N p(\bs c_n^{(k)} | A^{(k)}, \bs t)  \right\}
& \nonumber \\
&  \prod_{n=1}^N \prod_{\tau=1}^{L_n} p(t_{n,\tau} | \bs T_{t_{n,\tau-1}}) p(x_{n,\tau} | t_{n,\tau}, \bs\rho_{t_{n,\tau}}) & \nonumber \\
& \prod_{j=1}^J p(\bs T_j | \bs\gamma_j) p(\bs\rho_j | \bs\kappa_j)&
%\prod_{s=1}^S \bigg\{ p(\bs \theta^{(s)})  \nonumber \\
%& p(B^{(s)} | \bs\beta^{(s)}) \! \! \prod_{n=1}^N \!\! \left\{ p(\bs x | \bs d^{(s)}, \bs \theta^{(s)}) p(\bs d^{(s)} | B^{(s)}, \bs t)  \right\} \!\! \bigg\}, \nonumber
& \label{eq:joint}
\end{flalign}
where 
$\bs c$ is the set of annotations for all documents from all annotators,
$\bs t$ is the set of all sequence labels for all documents,
$N$ is the number of documents, 
$L_n$ is the length of the $n$th document, 
$J$ is the number of classes,
 $\bs x$ is the set of all word sequences for all documents and
$\bs\rho$, $\bs\gamma$ and $\bs\kappa$ are the sets of parameters for all
label classes.
 %=\{\bs c^{(1)}, .., \bs c^{(K)} \}$,
%
%Terms distribution omit subscripts and superscripts are the sets of  parameters for all values of the omitted index.

%\begin{flalign}
%& 
%=
%p\left(d_{n,\tau}^{(s)} | \bs t, d_{n,\tau-1}^{(s)}, A^{(s)} \right) & \nonumber\\
%& p \left(d_{n,\tau}^{(s)} | \bs\phi_n, \bs\theta^{(s)} \right) / p \left(d_{n,\tau}^{(s)} | \bs\theta^{(s)} \right),  & 
%\end{flalign}
%% Imagine what happens for each of the different combinations of values of d. We could train a squillion different sequence taggers on these, then take a weighted sum.
%
%% multiply by p(phi | theta) to get joint, divide by p(d) to get likelihood of phi given d, then since phi is independent of t and A,  p(phi | theta) cancels out. This follows from the generative model, which should be described here. The inference section should 
%% talk about learning theta, and computing the expectation of d.
%%We could avoid all this if the sequence tagger just learns to represent the left hand side. So t and A stay in the condition for the likelihood of the features. In which case, the first term on the right hand side is like a prior that needs to be
%% used when learning theta.
%where the first term on the right-hand side is defined by the annotator model
%with parameters $A^{(s)}$, and 
%of the sequence tagger, $s$.
%integrating existing sequence taggers using the learning
%procedure described in the next section.

\section{Inference using Variational Bayes} \label{sec:vb}
 
Given a set of annotations, $\bs c$, 
%=\{\bs c^{(1)}, .., \bs c^{(K)} \}$, from $K$ annotators,
we obtain a posterior distribution over 
%the parameters, 
%$\bs T$, $\bs\theta = \{ \bs\theta^{(1)},..,\bs\theta^{(S)} \}$, 
%$\bs A=\{ A^{(1)},..,A^{(K)} \}$ and
%$\bs B = \{ B^{(1)},..,B^{(S)} \}$, 
%and thereby compute the posterior over 
sequence labels, $\bs t$, using
\emph{variational Bayes} (\emph{VB})~\cite{attias_advances_2000}.
Unlike maximum likelihood methods such as standard expectation maximization (EM),
VB considers prior distributions 
and accounts for parameter uncertainty due to 
noisy or sparse data.
In comparison to other Bayesian approaches such as Markov chain Monte Carlo (MCMC),
VB is often faster, readily allows incremental learning, and provides easier ways
to determine convergence~\cite{Bishop2006}. 
It has been successfully applied to a wide range of methods,
including being used as the standard learning procedure for LDA~\cite{blei2003},
and to combining non-sequential crowdsourced classifications~\cite{simpsonlong}.

The trade-off is that we must approximate the posterior distribution with
%the \emph{mean field} assumption,
an approximation that
factorises between subsets of latent variables:
% so that each subset, $z$, has a \emph{variational factor}, $q(z)$:
%over $\bs t$ and the model parameters, 
%$\bs T$, $\bs\theta = \{ \bs\theta^{(1)},..,\bs\theta^{(S)} \}$, 
%$\bs A=\{ A^{(1)},..,A^{(K)} \}$ and
%$\bs B = \{ B^{(1)},..,B^{(S)} \}$.
%The labels produced by the sequence taggers, $\bs d$, 
%can be marginalized, so do not appear in the approximate posterior, which is given by:
\begin{flalign} \label{eq:vb_posterior}
& p(\bs t, \bs A, \bs T, \bs\rho | \bs c, \bs x, \bs \alpha, \bs\gamma, \bs\kappa ) &   \nonumber\\
&  \approx \prod_{k=1}^K  q(A^{(k)}) \prod_{j=1}^J \left\{
q(\bs T_j) q(\bs \rho_j)\right\} \prod_{n=1}^N q(\bs t_n) .
\end{flalign}
VB performs approximate inference by
%we optimize Equation \ref{eq:vb_posterior} 
%using coordinate ascent to 
updating each variational factor, $q(z)$, in turn,  
optimising the approximate posterior distribution until it converges.
%taking expectations with respect to the current estimates of the other variational factors.
%(see Algorithm \ref{al:vb_bac}).
%Each iteration reduces the KL-divergence between the true and approximate posteriors
%of Equation \ref{eq:vb_posterior}, and hence optimizes a lower bound on the 
%log marginal likelihood, also called the evidence lower bound or ELBO
Details of the theory are beyond the scope of this paper, but are  
explained by ~\citet{Bishop2006}.
The VB algorithm is described in Algorithm \ref{al:vb_bac},
making use of update equations for the variational factors given below.
% We now provide the variational factors,
% which can be used to approximate the marginal posterior distributions for the parameters and sequence
% labels,
% and explain how to incorporate existing sequence taggers into the algorithm.
%that
%predict only sequence labels rather than full posterior distributions.
\begin{algorithm}
\DontPrintSemicolon
 \KwIn{ Annotations $\bs c$, tokens $\bs x$}
 \nl Compute initial values of $\mathbb{E}\ln A^{(k)},\forall k$,
 $\mathbb{E}\ln \rho_j,\forall j$, 
 $\mathbb{E}\ln \bs T_j,\forall j$ 
 from their prior distributions.\;
 \While{$\mathrm{not\_converged}(r_{n,\tau,j},\forall n,\forall \tau, \forall j)$}
 {
 \nl Update $r_{j,n,\tau}$, $s_{t_{j,n,\tau\!-\!1}, t_{\iota,n,\tau}}$, $\!\forall j,\!\forall \tau,\!\forall n,\!\forall \iota$,
 %given $\bs c$, 
 %$\mathbb{E}\left[\ln \bs A \right]$, $\mathbb{E}\left[\ln \bs B \right]$ and $\mathbb{E}\left[\ln \bs T \right]$
 using forward-backward algorithm
 given $\bs x$, $\bs c$, $\mathbb{E}\ln \bs T_j,\forall j$,
 $\mathbb{E}\ln\bs\rho_j,\forall j$, 
  $\mathbb{E}\ln\bs A^{(k)},\forall k$.\;%~\cite{ghahramani2001introduction}\;
 %\nl Set current true label predictions $\mathbb{E}\left[t_{n,\tau}=j\right] = r_{j,n,\tau}$ \;
%  \nl Retrain all sequence taggers using $\tilde{\bs d}^{(s)}$ as training labels for 
%  each tagger $s$\;
%  \nl Use sequence taggers to predict $\hat{d}^{(s)}_{n,\tau}(i)$, $\forall s,\forall n,\forall \tau,\forall i$\;
 \nl Update $\mathbb{E}\ln A^{(k)},\forall k$, 
 given $\bs c$, $r_{j,n,\tau}$.\;
%  \nl Update $\ln q(B^{(s)})$ and $\mathbb{E}\ln B^{(s)},\forall s$ ,
%  given current $\hat{\bs d}$, $r_{j,n,\tau}$\;
 \nl Update $\mathbb{E}\ln T_{j,\iota},\forall j,\forall \iota$, 
 given $s_{t_{j,n,\tau\!-\!1}, t_{\iota,n,\tau}}$.\;
 \nl Update $\mathbb{E}\ln \rho_j,\forall j$ given $\bs x$, $r_{j,n,\tau}$.
 }
\KwOut{ Label posteriors, $r_{n,\tau,j},\!\forall n,\!\forall \tau, \!\forall j$, %\mathbb{E}[\bs t]$, 
most probable sequence of labels, $\hat{\bs t}_n,\forall n$ using Viterbi algorithm }
\;
\caption{The VB algorithm for BSC.}
\label{al:vb_bac}
\end{algorithm}

We compute the posterior probability of each true token label, 
$r_{n,\tau,j}= \mathbb{E}[p(t_{n,\tau}=j | \bs c)]$,  %_{\bs T,\bs A,\bs B,\bs d}
and of each label transition, $s_{n,\tau,j,\iota} = 
\mathbb{E}%_{\bs T,\bs A,\bs B,\bs d}
[p(t_{n,\tau\!-\!1}=j, t_{n,\tau}=\iota | \bs c)]$,
%
% terms provide the output predictions of the class labels. 
% To compute $q(\bs T_j)$, $q(\bs A^{(k)} )$, and $q(\bs B^{(s)})$, 
% we require expectations for the individual 
% true labels $r_{n,\tau,j} 
using the forward-backward algorithm~\cite{ghahramani2001introduction}.
In the forward pass, we compute:
\begin{flalign}
   & \ln r^{-}_{n,\tau,j} = \ln \sum_{\iota=1}^J \left\{ r^{-}_{n,\tau-1,\iota} \mathrm{e}^{\mathbb{E}\ln T_{\iota,j}} \right\} + ll_{n,\tau}(j), & \nonumber \\
% \end{flalign}
%  where the log likelihood $ll(j,n,\tau)$ of the annotations for token $\tau$ in document $n$ given
%  label $j$ is:
% \begin{flalign} 
   & ll_{n,\tau}(j) = \!\sum_{k=1}^K \!\mathbb{E}\!%_{\bs A}
   \ln A^{(k)}\!\!\left(j, c_{n,\tau}^{(k)}, c_{n,\tau\!-\!1}^{(k)} \right)
   \!\mathbb{E}\!\ln\rho_{j,x_{n,\tau}}, & 
   %+  \sum_{s=1}^S
%    & \nonumber \\
%    &  \sum_{i=1}^J\sum_{\iota=1}^J \mathbb{E}%_{\bs B}
%    \ln B^{(s)} \!\left(j, i, \iota \right)  
%    \hat{d}_{n,\tau}^{(s)}(i) \hat{d}_{n,\tau-1}^{(s)}(\iota), & 
 \end{flalign}
 and in the backward pass we compute:
  \begin{flalign}
  %& \ln \lambda_{n,L_n,j} = 0, \hspace{1cm}
   \ln \lambda_{n,\tau,j} & = \ln\sum_{\iota=1}^J \exp \big\{ 
   \ln \lambda_{i,\tau+1,\iota}
   & \nonumber \\
&  + \mathbb{E}\ln T_{j,\iota} + ll_{n,\tau+1}(\iota) \big\} .&
 \end{flalign}
 Then we can calculate the posteriors as follows:
 \begin{flalign}
 r_{n,\tau,j} &\propto r^{-}_{n,\tau,j}\lambda_{n,\tau,j}, &\\
 s_{n,\tau,j,\iota} &\propto  r^{-}_{n,\tau-1,j} \lambda_{n,\tau,\iota} \exp\{\mathbb{E}\ln T_{j,\iota}+ ll_{n,\tau}(\iota)\}. &
 \end{flalign}
 
%The variational factor $q(\bs t)$ requires the following expectation:
The expectations of $\ln \bs T$ and $\ln\bs\rho$
can be computed using standard equations for a Dirichlet distribution:
 \begin{flalign}
& \mathbb{E}\ln T_{j,\iota} = \Psi\!\left(N_{j,\iota} \!\!+ \gamma_{j,\iota}\right) 
 - \Psi\!\left(\sum_{i=1}^J (N_{j,i} \!\!+ \gamma_{j,i}) \!\right), & \\
 & \mathbb{E} \ln\rho_j = \Psi\!\left(o_{j,w} \!\!+ \kappa_{j,w}\right) 
 - \Psi\!\left(\sum_{v=1}^J (o_{j,v} \!\!+ \kappa_{j,v}) \!\right), &
\end{flalign}
 where $\Psi$ is the digamma function,
  $N_{j,\iota} = \sum_{n=1}^N \sum_{\tau=1}^{L_n}  s_{n,\tau,j,\iota}$ is the expected number of times that label $\iota$ follows label $j$,
 and $o_{j,w}$ is the expected number of times that word $w$
 occurs with sequence label $j$.
Similarly, for the \emph{seq} annotator model, the expectation terms are:
 \begin{flalign}
 \label{eq:elna}
 \mathbb{E}\ln A^{(k)}(j,l,m) &= \Psi\!\left(N^{(k)}_{j,l,m}\right)  & \\
& - \Psi\left(\sum_{\;m'\!=1}^J \left( N^{(k)}_{j,l,m'} \right) \right). & \nonumber \\
N^{(k)}_{j,l,m} =  \alpha_{j,l,m}^{(k)} + & \sum_{n=1}^N \sum_{\tau=1}^{L_n} 
r_{n,\tau,j} \delta_{l,c^{(k)}_{n,\tau-1}}\delta_{m,c^{(k)}_{n, \tau}}, & 
 \end{flalign}
 where $\delta$ is the Kronecker delta. 
For other annotator models, this equation is simplified as the values of
the previous labels $c^{(k)}_{n,\tau-1}$ are ignored.
\subsection{Most Likely Sequence Labels}

%Two types of output from the BSC inference algorithm are of particular interest: (1) 
The approximate posterior probabilities of the true labels, $r_{j,n,\tau}$, provide confidence estimates for the labels. However, it is often useful to  compute 
the most probable sequence of labels, $\hat{\bs t}_n$, using the
Viterbi algorithm~\cite{viterbi1967error}. 
%To apply the algorithm, we use the converged variational factors to compute 
%$\mathbb{E}[\bs T]$ and $\mathbb{E}[A^{(k)}],\forall k$. %$\mathbb{E}[B^{(s)}],\forall s$ and
% $\hat{d}_{n,\tau}^{(s)}(i), \forall s, \forall n, \forall \tau, \forall i$.
The most probable sequence is particularly useful because, unlike $r_{j,n,\tau}$,
the sequence will be consistent with any transition 
constraints imposed by the priors on the transition matrix $\bs T$, 
such as preventing `O'$\rightarrow$`I' transitions by assigning them zero probability.

\section{Experiments}\label{sec:expts_all}

\begin{table}[t]
\small
\centering
\setlength{\tabcolsep}{3pt}
\begin{tabular}{l l l l l l l l} \toprule
data & \multicolumn{3}{l}{\#sentences or docs} & %#tokens & 
\multicolumn{2}{l}{\#annotators} & \#gold & span \\ %\multicolumn{2}{l}{span length}  \\
-set & total & dev & test
& %/sent. & 
total & /doc & spans & length \\ %mean & std.  \\
\toprule
NER & 6,056 & 2,800 & 3,256 & %13 & 
47 & 4.9 & 21,612 % mean per type (std) 5403 (1376) 
& 1.51 \\%& 0.75 \\
PICO & 9,480 & 191 & 191 & %150 & 
312 & 6.0 & 700 & 7.74 \\%& 7.38  \\
ARG & 8,000 & 60 & 100 & 105 & 5 & 73 &  17.52 \\
 \bottomrule
\end{tabular}
\caption{
%Numbers of sentences, annotators, and spans for datasets used in our experiments. Sentences with crowd all have crowdsourced labels. Only dev and test sentences have gold sequence labels.
Dataset statistics. Span lengths are means.
}
\label{tab:datasets}
\end{table}
 
% TODO: write about the hyper-parameters?
\begin{table*}
\small
\centering
\nprounddigits{1}
\npdecimalsign{.}
\setlength{\tabcolsep}{4pt}
\begin{tabular}{l l l l r@{\hskip 0.8cm} l l l r@{\hskip 0.8cm} l l l r }
\toprule
& \multicolumn{4}{l}{NER} & \multicolumn{4}{l}{PICO} & \multicolumn{3}{l}{ARG} \\ 
%& \multicolumn{4}{l}{ARG: relaxed} \\
& \text{Prec.} &  \text{Rec.} & \text{F1} & \text{CEE} & 
\text{Prec.} & \text{Rec.} & \text{F1} & \text{CEE} & 
%\text{Prec.} & \text{Rec.} & \text{F1} &
\text{Prec.} & \text{Rec.} & \text{F1} & \text{CEE} 
\\ \toprule
Best worker & 
76.4 & 60.1 & 67.3 & 17.1 & 
64.8 & 53.2 & 58.5 & 17.0 & 
62.7 & 57.5 & 60.0 & %79.2 & 76.4 & 77.8 & 
44.20 
\\
Worst worker & 
55.7 & 26.5 & 35.9 & 31.9 &
50.7 & 52.9 & 51.7 & 41.0 & 
25.5 & 19.2 & 21.9 & %80.0 & 41.3 & 54.5 & 
70.33
\\ \midrule
MV & 
79.9 & 55.3 & 65.4 & 6.24 &
82.5 & 52.8 & 64.3 & 2.55 & 
40.0 & 31.5 & 34.8 & %88.2 & 60.1 & 71.5 & 
14.03
\\ 
MACE & 
74.4 & 66.0 & 70.0 & 1.01 & 
25.4 & 84.1 & 39.0 & 58.2 & 
31.2 & 32.9 & 32.0 & %81.0 & 60.8 & 69.4 & 
2.62
\\ 
DS & 
79.0 & 70.4 & 74.4 & 2.80 & 
71.3 & 66.3 & 68.7 & 0.44 & 
% new implementation -- bad init? 34.6 & 75.1 & 47.4 & 0.40 &
45.6 & 49.3 & 47.4 & %80.3 & 76.4 & 78.3 & 
0.97
\\ 
IBCC & 
79.0 & 70.4 & 74.4 & \textbf{0.49} & 
72.1 & 66.0 & 68.9 & \textbf{0.27} &
44.9 & 47.9 & 46.4 & %80.3 & 76.4 & 78.3 & 
\textbf{0.85}
\\ \midrule
HMM-crowd & 
80.1 & 69.2 & 74.2 & 1.00 & 
75.9 & 66.7 & 71.0 & 0.99 &
43.5 & 37.0 & 40.0 & %87.0 & 61.8 & 72.2 & 
3.38
\\ 
\midrule
BSC-acc & 
\textbf{83.4} & 54.3 & 65.7 & 0.96 &
\textbf{89.4} & 45.2 & 60.0 & 1.59 &  
36.9 & 32.9 & 34.8 & %87.7 & 59.2 & 70.7 & 
6.47
\\ 
BSC-spam &
67.9 & 74.1 & 70.9 & 0.89 &
46.7 & \textbf{84.4} & 60.1 & 1.98 & 
55.7 & 53.4 & 54.5 & %85.8 & 77.3 & 81.3 & 
2.80
\\ 
BSC-CV & 
83.0 & 64.6 & 72.6 & 0.93 &
74.9 & 67.2 & 71.1 & 0.84 & 
37.9 & 34.2 & 36.0 & %87.2 & 60.5 & 71.5 &
 4.73
 \\ 
BSC-CM & 
79.9 & 72.2 & 75.8 & 1.46 & 
60.1 & 78.8 & 68.2 & 1.49 & 
\textbf{56.0} & 57.5 & 56.8 & %80.3 & 79.1 & 79.7 & 
3.76 
\\ 
BSC-seq & 
80.3 & \textbf{74.8} & \textbf{77.4} & 0.65 & 
70.4 & 75.3 & \textbf{72.8} & 0.53 & 
54.4 & \textbf{67.1} & \textbf{60.1} & %70.2 & 86.5 & 77.5 &
 3.26
\\ \midrule
BSC-CM-notext & 
74.7 & 69.7 & 72.1 & 1.48 & 
62.7 & 74.8 & 68.2 & 1.32 & 
55.1 & 58.9 & 57.0 & 2.75
\\
BSC-CM$\backslash\bs T$ & 
80.0 & 73.0 & 76.3 & 0.99 &
65.8 & 66.7 & 66.2 & 0.28 &
52.9 & 49.3 & 51.1 & 1.69
 \\
BSC-seq-notext & 
81.3 & 71.9 & 76.3 & 0.52 & 
81.2 & 59.2 & 68.5 & 0.73 &
36.9 & 52.0 & 43.2 & 5.64 
\\ 
BSC-seq$\backslash\bs T$ & 
64.2 & 44.4 & 52.5 & 0.77 &
51.2 & 70.4 & 59.8 & 1.04 &
0.11 & 0.05 & 0.07 & 6.38
\\
\bottomrule 
\end{tabular}
\caption{Aggregating crowdsourced labels: estimating true labels for documents labelled by the crowd.}
\label{tab:aggregation_results}
\npnoround
\end{table*}

\paragraph{Datasets. }\label{sec:expts}

We compare BSC to alternative methods on three NLP datasets
%to test performance in passive and active learning scenarios, 
%analyze errors, and
%visualize the learned annotator models.
containing both crowdsourced and gold sequential annotations.
\emph{NER}, the CoNLL 2003 named-entity recognition dataset~\cite{tjong2003introduction},
which contains gold labels for four named entity categories (PER, LOC, ORG, MISC),
with crowdsourced labels provided by \cite{rodrigues2014sequence}.
\emph{PICO}~\cite{nguyen2017aggregating}, 
consists of medical paper abstracts that have been annotated by a crowd to indicate text spans that identify the population enrolled in a clinical trial. 
\emph{ARG}~\cite{trautmann2019robust} contains a mixture of argumentative and non-argumentative sentences, in which the task is to mark the spans 
that contain pro or con arguments for a given topic. 
Dataset statistics are shown in Table \ref{tab:datasets}. 
The datasets differ in typical span length, with argument components in ARG the longest, while named entities in NER spans are often only
one token long.

The gold-labelled documents are split into validation and test sets. 
For NER, we use the split given by Nguyen et al.~\shortcite{nguyen2017aggregating},
while for PICO and ARG, we make random splits since the splits from previous work
were not available\footnote{The data splits are available from our Github repository.
Since we use different splits, our results for PICO are not identical to ~\citet{nguyen2017aggregating}.}.

\paragraph{Evaluation metrics. }
For NER and ARG we use the CoNLL 2003 F1-score, which considers only exact span matches %(type, start and end) 
to be correct. Incomplete named entities are typically not useful, and for ARG, it is desirable to identify complete argumentative units that
make sense on their own. 
%This measure is intuitive because complete named entities must be marked to be of value. 
For medical trial populations, partial matches still contain useful information, so for PICO we use a relaxed F1-score, as in ~\citet{nguyen2017aggregating}, 
which counts the matching fractions of spans when computing precision and recall. 
%Since the spans in PICO are longer than those of NER, 
 
%We additionally compute the root mean squared error in the span lengths, i.e. the difference between the  % actually it's not quite that. We computed the difference in mean span lengths. This is already captured by F1 score for pico and better described by the span-level-precision and recall. Our metric might be more if we didn't take the absolute so we could see if spans were often too long or too short.
We also compute the cross entropy error (\emph{CEE}, also known as log-loss).
While this is a token-level rather than span-level metric, it evaluates the quality of the probability estimates produced by aggregation methods, which are useful for tasks such as active learning, as it penalises over-confident mistakes more heavily.

\paragraph{Evaluated Methods. }
We evaluate BSC in combination with all of the annotator models described in Section \ref{sec:model}.
%to assess whether the sequential annotator model, \emph{seq},
%improves the quality of the inferred sequence tags. 
As well-established non-sequential baselines, we include token-level majority voting (\emph{MV}), 
\emph{MACE}~\cite{hovy2013learning} which uses the \emph{spam} annotator
model,
Dawid-Skene (\emph{DS})~\cite{dawid_maximum_1979}, which uses the \emph{CM} annotator model,
 and independent Bayesian classifier combination (\emph{IBCC})~\cite{kim2012bayesian}, which is a Bayesian treatment of Dawid-Skene. 
We also compare against the state-of-the-art sequential \emph{HMM-crowd} method \cite{nguyen2017aggregating}, which uses a combination of 
maximum \emph{a posteriori} (or smoothed maximum likelihood) estimates for the \emph{CV} annotator model 
and variational inference for an integrated hidden Markov model (HMM). 
%  HMM-crowd is the current state-of-the-art and allows us to compare our approach against 
%  a model without a fully Bayesian treatment. 
%We also introduce a \emph{clustering} baseline,
%that aggregates spans from multiple annotators by grouping them together
%using kernel density estimation~\cite{rosenblatt1956remarks}.
%MACE and IBCC are variants of BSC-spam and BSC-CM, respectively, with non-sequential true label models.
%and serve to show the benefits of the sequential BSC model.
HMM-Crowd and DS use non-Bayesian inference steps and can be compared with
their Bayesian variants, BSC-CV and IBCC, respectively. 

Besides the annotator models, BSC also makes use of text features and a transition matrix, $\bs T$, over true labels.
We test the effect of these components by running BSC-CM and BSC-seq with no text features (\emph{notext}), 
and without the transition matrix, which is
 replaced by simple independent class probabilities (labelled $\backslash \bs T$).
 
We tune the hyperparameters using grid search on the development sets. To limit the number of hyperparameters to tune, 
we optimize only three values for BSC:
hyperparameters of the transition matrix, $\bs\gamma_j$, are set to the same value, 
$\gamma_0$, except for disallowed transitions, (O$\Arrow{0.2cm}$I and transitions between types, e.g. I-PER$\Arrow{0.2cm}$I-ORG), 
which are set to $1e-6$; 
for the annotator models,
all values are set to $\alpha_0$, except for disallowed transitions, which are set to $1e^{-6}$,
then $\epsilon_0$ is added to hyperparameters 
corresponding to correct annotations (e.g. diagonal entries in a confusion matrix).
This encodes the prior assumption that annotators are more likely to have an accuracy greater than random,
which avoids the non-identifiability problem in which the class labels are switched around.
%We use validation set F1-scores to choose values from $[0.1, 1, 10, 100]$, 
%training on a small subset of 250 documents for NER and 350 documents for PICO. 

\begin{table*}[h]
\small
\centering
\setlength{\tabcolsep}{4pt}
\begin{tabular}{l l l l l l l l l l l l l l l }
\toprule
Method & Data & exact & wrong & partial  & missed  & false & late & early & late & early & fused & splits & inv- &  length \\ 
 & -set & match & type & match & span & +ve & start & start & finish & finish & spans &  & alid & error \\
\midrule
MV & NER & 2869 & 304 & 196 & 1775 & 100 & 96 & 10 & 15 & 85 & 17 & 26 & 81 & 0.04 \\
IBCC & NER & 3742 & 386 & 187 & 829 & 345 & 107 & 27 & 43 & 77 & 47 & 29 & 74 & 0.12 \\
HMM-crowd & NER & 3650 & 334 & 115 & 1045 & 210 & 109 & 22 & 33 & 89 & 37 & 23 & 0 & 0.03 \\
BSC-CV & NER & 3381 & 284 & 80 & 1399 & 121 & 94 & 17 & 18 & 90 & 22 & 8 & 0 & 0.00 \\
BSC-CM & NER & 3856 & 362 & 63 & 863 & 315 & 124 & 25 & 63 & 77 & 53 & 13 & 0 & 0.14 \\
BSC-seq & NER & 3995 & 353 & 110 & 686 & 357 &  84 &  29 &  25 &  88 &  28 &  26 & 0 & 0.09 \\
\midrule 
MV & PICO  & 144 & 0 & 60 & 145 & 48 & 9 & 11 & 1 & 0 & 3 & 9 & 40 & 1.26 \\
IBCC & PICO & 193 & 0 & 53 & 103 & 100 & 14 & 10 & 0 & 2 & 3 & 10 & 19 & 0.45 \\
HMM-crowd& PICO & 189 & 0 & 54 & 106 & 84 & 13 & 21 & 0 & 0 & 5 & 8 & 0 & 1.99 \\
BSC-CV     & PICO & 156 & 0 & 76 & 117 & 81 & 10 & 25 & 0 & 0 & 11 & 0 & 0 & 2.15 \\
BSC-CM     & PICO & 216 & 0 & 50 & 83 & 157 & 10 & 19 & 0 & 0 & 4 & 17 & 0 & 2.42\\
%174 & 0 & 98 & 192 & 18 & 15 & 8  & 0 & 4 & 18 \\
BSC-seq    & PICO & 168 & 0 & 86 & 95 & 67 & 17 & 19 & 5 & 0 & 4 & 9 & 0 & 0.61 \\
\midrule 
MV & ARG & 17 &  0 & 26 & 14 &  4  &  9 & 1 &  0 & 2 &  0 &  0 & 9 & 5.27 \\
IBCC & ARG & 27 & 1 & 21 & 8 & 9 & 7 & 2 & 0 & 1 & 0 & 3 & 9 & 3.43 \\
HMM-Crowd & ARG & 20 & 0 & 23 & 14 & 4 & 7 & 2 & 0 & 2 & 0 & 0 & 4 & 4.87 \\
BSC-CV & ARG & 18 & 0 & 25 & 14 & 4 & 12 & 2 & 0 & 2 & 0 & 0 & 0 & 5.37 \\
BSC-CM & ARG & 35 & 1 & 12 & 9 & 9 & 7 & 2 & 0 & 1 & 1 & 0 & 0 & 2.11 \\
BSC-Seq & ARG & 39 & 3 & 12 & 3 & 20 & 6 & 4 & 0 & 0 & 1 & 0 & 0 & 0.46 \\
\bottomrule
\end{tabular}
\caption{Counts of different types of span errors.}
\label{tab:error_analysis}
\end{table*}

\paragraph{Aggregation Results. }\label{sec:task1}

This task is to combine multiple crowdsourced labels and predict the true labels.
%For both datasets, we provide all the crowdsourced labels as input to the aggregation method. 
The results are shown in Table \ref{tab:aggregation_results}.
%Although DS and IBCC do not consider sequence information nor the text itself, 
%they both perform well on both datasets,
%with IBCC reaching better cross entropy error than DS due to its Bayesian treatment.
%against HMM-crowd on NER,
%and BSC-CM variants on PICO. 
%The improvement of DS over the results given 
%by Nguyen et al. ~\shortcite{nguyen2017aggregating} may be due to implementation differences. 
%Neither MACE, BSC-acc nor BSC-spam perform strongly, with F1-scores sometimes falling below MV. 
%The acc and MACE annotator models may be a poor match for the sequence labelling task if annotator
%competence varies greatly depending on the true class label.
%The annotator models of BSC-CV and BSC-CM are better, although BSC-CM performs worse on PICO.
BSC-seq outperforms the other approaches,
including the previous state of the art, HMM-crowd (significant on all datasets
 with $p\ll.01$ using a two-tailed Wilcoxon signed-rank test).
%despite
%having a larger number of parameters to learn.
Without the text model (BSC-seq-notext) or the transition matrix (BSC-seq$\backslash\bs T$),
BSC-seq performance decreases heavily,
while BSC-CM-notext and BSC-CM$\backslash\bs T$ in some cases outperform
BSC-CM.
This suggests that $\emph{seq}$, with its greater number of parameters to learn, 
is most effective in combination with the transition matrix and simple text model.
On the ARG dataset, the scores are close to zero for BSC-seq$\backslash\bs T$.
Further investigation shows that this is because 
BSC-seq$\backslash\bs T$ identifies many spans with one 
or two incorrect labels. Since we use exact span matches to compute true
and false positives, these small errors reduce the scores substantially.
In particular, we find a large number of missing ‘B’ tags at the start of spans
and misplaced ‘O’ tags that split spans in the middle.

The performance of all methods across the three datasets varies greatly.
With NER, the spans are short and the task is less subjective than PICO or ARG,
hence its higher F1 scores. PICO uses a relaxed F1 score, meaning its scores are
only slightly lower despite being a more ambiguous task. The constitution of an 
argument is also ambiguous, so ARG scores are lower, particularly as we use  
strict span-matching to compute the F1 scores. Raising the scores may be possible in future
by using expert labels as training data, i.e. as known values of $\bs t$,
which would help to put more weight on annotators with similar labelling 
patterns to the experts.

We categorise the errors made by key methods and list the
counts for each category in Table \ref{tab:error_analysis}.
All machine learning methods tested here reduce the number of spans that were completely missed by majority voting. Note that BSC completely removes all ``invalid'' spans
(O$\Arrow{0.2cm}$I transitions) due to the sequential model with prior hyperparameters set
to zero for those transitions.
For PICO and ARG, which contain longer spans,
BSC-seq has lower ``length error'' than other methods,
 which is the mean difference in number of tokens between the predicted and gold spans. 
 It also reduces the number of missing spans, although in NER and ARG that comes at the cost of producing more false positives (predicting spans where there are none).
Overall, BSC-seq appears to be the best choice for identifying 
exact span matches and reducing missed spans.

\begin{figure*}[h]
\begin{minipage}[b][0.5cm][b]{0.1\textwidth} 
Previous label = I:\\
\vspace{1cm}
\end{minipage}
  \includegraphics[width=0.88\textwidth, clip=True, trim=10 20 10 28]{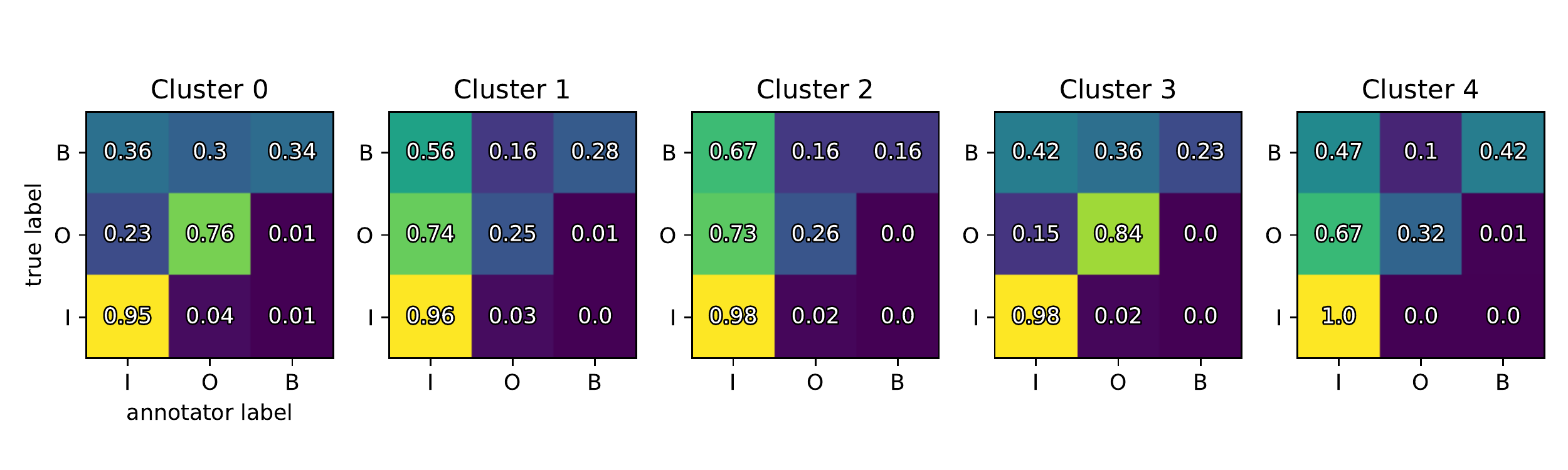}
\vspace{0.2cm}
\\
\begin{minipage}[b][0.5cm][b]{0.1\textwidth} 
Previous label = O:\\
\vspace{1cm}
\end{minipage}
  \includegraphics[width=0.88\textwidth, clip=True, trim=10 20 10 50]{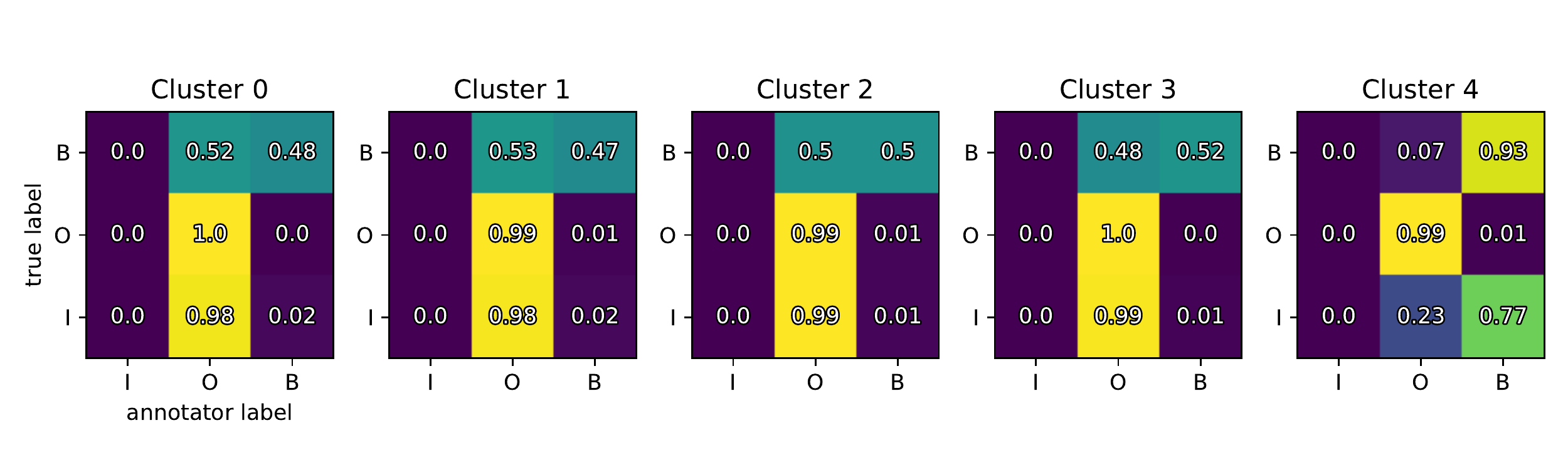}
\vspace{0.2cm}
\\
\begin{minipage}[b][0.5cm][b]{0.1\textwidth} 
Previous label = B:\\
\vspace{1cm}
\end{minipage}
  \includegraphics[width=0.88\textwidth, clip=True, trim=10 20 10 50]{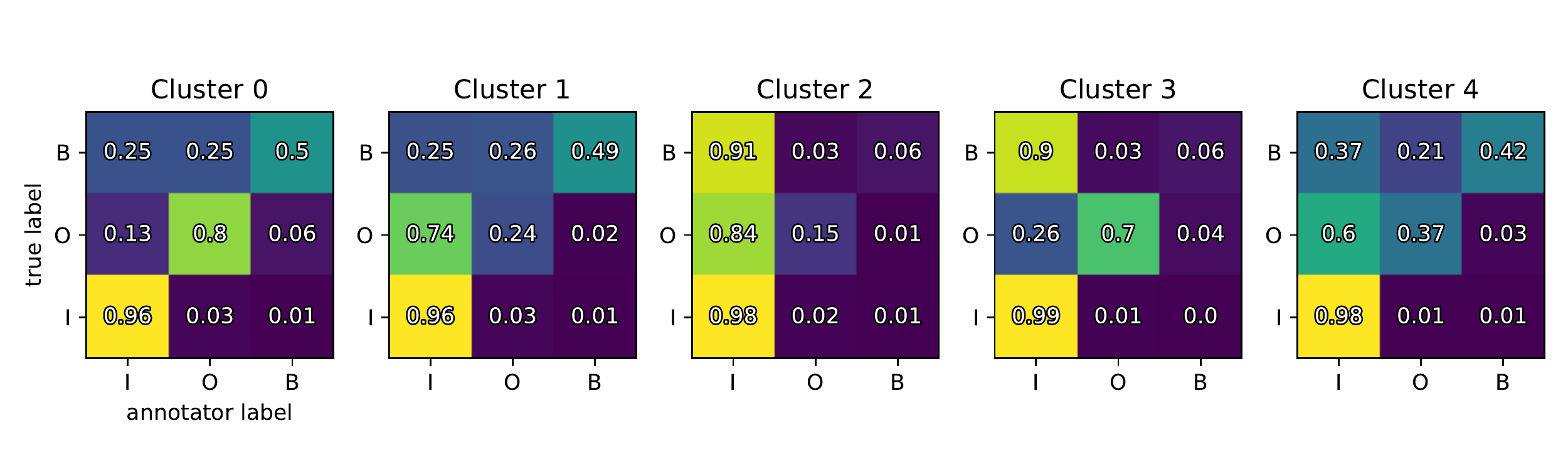}
\\
\centering\includegraphics[width=0.45\textwidth,clip=True,trim=0 40 110 260]{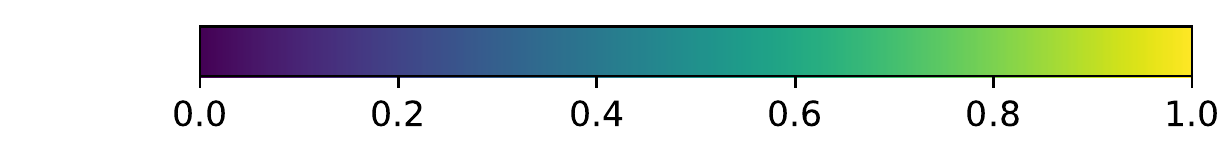}\\
\caption{Clusters of annotators in the PICO dataset. Each column shows the mean of the confusion matrices in the 
\textit{seq} annotator model for the members of one cluster, as inferred by BSC-seq.
 Each row corresponds to the mean confusion matrices conditioned on the annotator's previous label. 
 The confusion matrices are plotted as heatmaps,
where colours indicate the probabilities, which are also given by the numbers in each square.
}
\label{fig:conf_mat_clusters}
\end{figure*}
% TODO: make this plot more compact and relevant by plotting only the means across the whole dataset, 
%and the distribution of hellinger distances between the confusion matrices for each previous label case. Put on a single row.
%Table \ref{tab:aggregation_results} shows a benefit of using the sequential annotator model over CM, CV and acc.
\paragraph{Visualising Annotator Models. }
To determine whether, in practice, BSC-seq really learns distinctive confusion matrices depending on the previous labels,
we plot the learned annotator models for
PICO as probabilistic confusion matrices in Figure \ref{fig:conf_mat_clusters} (for an alternative visualisation, 
see Figure \ref{fig:anno_models_2} in the appendix).
%As the dataset contains a large number of annotators, we 
We clustered 
the confusion matrices % inferred by each model
into five groups by applying K-means to their posterior expected values,
then plotted the means for each cluster.
Each column in the figure shows the confusion matrices corresponding to the same cluster of annotators. 
In all clusters, BSC-seq learns different 
%accuracies for
confusion matrices depending on the previous label.
% B, I and O (the diagonal entries),
%which suggests that the data supports the use of this more complex model.
% explain ho
%These differences may explain its
%improvement over BSC-acc.
%BSC-CM differs from BSC-CV in that %has more distinctive clusters and 
%the first, fourth and fifth clusters 
%have off-diagonal values with different heights for the same true label value.
% The second 
%cluster for BSC-CM encodes likely spammers who usually choose 'O' regardless of the 
%ground truth. 
%Unlike BSC-CM, BSC-seq improved performance on PICO over BSC-CV. 
%The confusion matrices for BSC-seq are
%very different depending on the worker's previous annotation. 
There are also large variations between the clusters.
The third column, for example, shows
annotators with a tendency toward B$\Arrow{0.2cm}$I transitions regardless of the true label, while other clusters 
indicate very different labelling behaviour. The model therefore appears able to learn
distinct confusion matrices for different workers given previous labels, 
which supports the use of sequential
annotator models.

 \paragraph{Active Learning. }
 \begin{figure}[h]
 \centering
   \includegraphics[width=0.3\columnwidth, clip=True, trim=530 160 20 20]{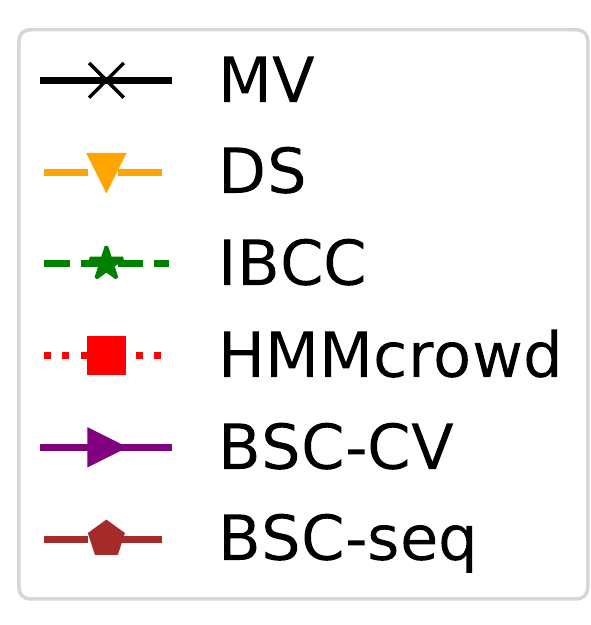}
      \includegraphics[width=0.9\columnwidth, clip=True, trim=40 22 190 15]{figures/NER_AL/pool/plot_F1-score.pdf}
 \caption{F1 scores for active learning simulations on NER using least-confidence 
 uncertainty sampling.
 %: prediction performance after each labelled batch is received. Mean scores over 10 repeats.
 }
 \label{fig:alner}
 \end{figure}
 Active learning is an iterative process that can reduce the amount of labelled
  data required to train a model.
  At each iteration, the active learning strategy selects informative data points,
  obtains their labels, 
  then re-trains a labelling model given the new labels.
 The updated model is then used in the next iteration to identify the most
 informative data points.  
 
  We simulate active learning in a crowdsourcing scenario
  where the goal is to learn the true labels,
  $\bs t$, by selecting documents for the crowd to label.
  Each document can be labelled multiple times by different workers.
  In contrast, in a passive learning setup, the number of annotations per document is 
  usually constant across all documents. 
  For example, in the PICO dataset, each
  document was labelled six times.
  The aim of active learning is to decrease the number of annotations required
  by avoiding relabelling documents whose true labels can be determined with  high confidence from fewer labels.
  
  We simulate active learning using 
 %a well-established technique, \emph{uncertainty sampling}
 the \emph{least confidence} strategy, shown to be effective by 
 ~\citet{settles2008analysis},
 as described in Algorithm \ref{al:uncertainty_sampling}.
At each iteration, we estimate $\bs t$ from the current set of 
  crowdsourced labels, $\bs c$,
  using one of the methods from 
  our previous experiments as the labelling model,
  then use this model to determine the least confident $batch\_size$ 
 documents to be labelled by the crowd. 
  If the simulation has requested all the labels for a document that
  are available in our dataset, this document is simply ignored when 
  choosing new batches
  and is not selected again.
 
 We hypothesise that BSC will learn more quickly than non-sequential 
 and non-Bayesian methods
 in the active learning simulation. 
 Bayesian methods such as BSC account for uncertainty in the model parameters when estimating confidence,  
 hence can be used to choose data points that rapidly improve
 the model itself.
 Sequential models such as BSC can also account for dependencies between sequence labels
 when estimating confidence.
% From the reviews:
%We apologise that the active learning experiment was not clear from the description. We use active learning to learn the aggregation models themselves, i.e. to train BSC-seq, BSC-CV, MV, etc. The F1-scores therefore evaluate the outputs of the aggregation models at each active learning round. The aim is to show how well each aggregation method performs on smaller numbers of annotations chosen by active learning. BSC-seq performs strongest, suggesting that using it with active learning could reduce the number of labels you need to obtain a desired F1 score for your aggregated labels. 
 %In contrast, frequentist methods
 %such as maximum likelihood output probabilities that do not account for parameter uncertainty due to 
 %small datasets or noisy labels. 
 
 %While various active learning methods could be applied here, in this paper we wish
 %to demonstrate only that BSC may serve as a good foundation for active learning,
 %and defer a deeper investigation of active learning techniques to future work.
 %TODO does there need to be another index in t for the document?
 \begin{algorithm}
 \DontPrintSemicolon
  \KwIn{ A random $initial\_set$ of training labels, the same for all methods. }
  \nl Set training set $\bs c=initial\_set$ \;
  \While{training set size $<max\_no\_labels$}
  {
  \nl Train model on $\bs c$ \;
  \nl For each document $n$, 
  compute $LC_n = 1-p(\bs t_n^*|\bs c)$, where $\bs t_n^*$ is the probability of the most likely sequence of labels 
  for $n$. \;
  \nl Obtain annotations for $batch\_size$ documents with the highest values of $LC$ (least confidence), and add them to $\bs c$\;
  }
 \caption{Active learning simulation using least-confidence sampling.}
 \label{al:uncertainty_sampling}
 \end{algorithm}
 
 Figure \ref{fig:alner} plots the mean F1 scores over ten repeats of the active learning simulation on the NER dataset (for clarity, we only plot key methods).
 When the number of iterations is very small, neither IBCC nor DS are able to outperform majority vote, and only produce a very small
 benefit as the number of labels grows. This highlights the need for a sequential model such as BSC or HMM-crowd for
 effective active learning with small numbers of labels.
 IBCC learns slightly quicker than DS,
 while BSC-CV clearly outperforms HMM-crowd: we believe this difference is due to the Bayesian treatment of IBCC and BSC,
 which means they are better able to estimate confidence than DS and HMM-crowd, which use maximum likelihood and maximum a posteriori inference.
 BSC-seq produces the best overall performance, and the gap grows as the number of labels increases, 
 since more data is required to learn the more complex annotator model.  

\section{Conclusions}

We proposed BSC, a novel Bayesian approach to aggregating sequence labels
that can be combined with several different models of annotator noise and bias.
To model the effect of dependencies between labels on annotator noise and bias, we introduced 
the \emph{seq} annotator model.
Our results demonstrated the benefits of BSC over established non-sequential methods, such 
as MACE, Dawid-Skene (DS), and IBCC.
%reinforce previous work that has demonstrated the benefits of modeling annotator reliability when aggregating noisy data, such as crowdsourced labels. 
We also showed the advantages of a Bayesian approach for active learning,
and that the combination of \emph{BSC} with the \emph{seq} annotator model improves 
the state-of-the-art over HMM-crowd on three NLP tasks with different types of span annotations.

In future work, we plan to adapt active learning methods for easy deployment on crowdsourcing platforms,
and to investigate techniques for automatically selecting good hyperparameters without recourse to a development
set, which is often unavailable at the start of a crowdsourcing process.

\section*{Acknowledgments}

This work was supported by the German Research Foundation through the German-Israeli Project Cooperation (DIP, grant DA 1600/1-1 and grant GU 798/17-1), and 
by the German Research Foundation (DFG) as part of the QA-EduInf project (grant GU 798/18-1 and grant RI 803/12-1). 
We would like to thank all of those who developed the datasets used
in this work.

% \addcontentsline{toc}{chapter}{Bibliography}
%\bibliographystyle{apalike}
% \bibliographystyle{IEEEtran}
\bibliographystyle{acl_natbib}
\bibliography{simpson}

\begin{thebibliography}{32}
\expandafter\ifx\csname natexlab\endcsname\relax\def\natexlab#1{#1}\fi

\bibitem[{Attias(2000)}]{attias_advances_2000}
Hagai Attias. 2000.
\newblock \href {http://www.gatsby.ucl.ac.uk/publications/papers/03-2000.pdf}
  {A variational {Bayesian} framework for graphical models}.
\newblock In \emph{Advances in Neural Information Processing Systems 12}, pages
  209--215. MIT Press.

\bibitem[{Bachrach et~al.(2012)Bachrach, Minka, Guiver, and
  Graepel}]{bachrach2012grade}
Yoram Bachrach, Tom Minka, John Guiver, and Thore Graepel. 2012.
\newblock \href {https://icml.cc/2012/papers/597.pdf} {How to grade a test
  without knowing the answers: a {Bayesian} graphical model for adaptive
  crowdsourcing and aptitude testing}.
\newblock In \emph{Proceedings of the 29th International Coference on
  International Conference on Machine Learning}, pages 819--826. Omnipress.

\bibitem[{Bishop(2006)}]{Bishop2006}
Christopher.~M. Bishop. 2006.
\newblock \href {http://cds.cern.ch/record/998831} {\emph{Pattern recognition
  and machine learning}}, 4th edition.
\newblock Information Science and Statistics. Springer.

\bibitem[{Blei et~al.(2003)Blei, Ng, and Jordan}]{blei2003}
David~M. Blei, Andrew~Y. Ng, and Michael~I. Jordan. 2003.
\newblock \href {http://www.jmlr.org/papers/v3/blei03a.html} {Latent
  {Dirichlet} allocation}.
\newblock \emph{The Journal of Machine Learning Research}, 3:993--1022.

\bibitem[{Dawid and Skene(1979)}]{dawid_maximum_1979}
Alexander~Philip Dawid and Allan~M. Skene. 1979.
\newblock \href {http://www.jstor.org/stable/2346806} {Maximum likelihood
  estimation of observer error-rates using the {EM} algorithm}.
\newblock \emph{Journal of the Royal Statistical Society. Series C {(Applied}
  Statistics)}, 28(1):20--28.

\bibitem[{Dietterich(2000)}]{dietterich2000ensemble}
Thomas~G Dietterich. 2000.
\newblock \href {https://link.springer.com/chapter/10.1007/3-540-45014-9_1}
  {Ensemble methods in {Machine Learning}}.
\newblock In \emph{Multiple classifier systems}, pages 1--15. Springer.

\bibitem[{Donmez et~al.(2010)Donmez, Carbonell, and
  Schneider}]{donmez2010probabilistic}
Pinar Donmez, Jaime Carbonell, and Jeff Schneider. 2010.
\newblock \href {https://epubs.siam.org/doi/abs/10.1137/1.9781611972801.72} {A
  probabilistic framework to learn from multiple annotators with time-varying
  accuracy}.
\newblock In \emph{Proceedings of the 2010 SIAM International Conference on
  Data Mining}, pages 826--837. SIAM.

\bibitem[{Felt et~al.(2016)Felt, Ringger, and Seppi}]{Felt2016SemanticAA}
Paul Felt, Eric~K. Ringger, and Kevin~D. Seppi. 2016.
\newblock \href {https://www.aclweb.org/anthology/C16-1168/} {Semantic
  annotation aggregation with conditional crowdsourcing models and word
  embeddings}.
\newblock In \emph{International Conference on Computational Linguistics},
  pages 1787--1796.

\bibitem[{Ghahramani(2001)}]{ghahramani2001introduction}
Zoubin Ghahramani. 2001.
\newblock \href
  {https://www.worldscientific.com/doi/abs/10.1142/9789812797605_0002} {An
  introduction to hidden markov models and {Bayesian} networks}.
\newblock \emph{International Journal of Pattern Recognition and Artificial
  Intelligence}, 15(01):9--42.

\bibitem[{Hovy et~al.(2013)Hovy, Berg-Kirkpatrick, Vaswani, and
  Hovy}]{hovy2013learning}
Dirk Hovy, Taylor Berg-Kirkpatrick, Ashish Vaswani, and Eduard~H Hovy. 2013.
\newblock \href {https://www.aclweb.org/anthology/N13-1132} {Learning whom to
  trust with {MACE}.}
\newblock In \emph{HLT-NAACL}, pages 1120--1130.

\bibitem[{Kim and Ghahramani(2012)}]{kim2012bayesian}
Hyun-chul Kim and Zoubin Ghahramani. 2012.
\newblock \href {http://www.jmlr.org/proceedings/papers/v22/kim12/kim12.pdf}
  {Bayesian classifier combination}.
\newblock In \emph{International Conference on Artificial Intelligence and
  Statistics}, pages 619--627.

\bibitem[{Lample et~al.(2016)Lample, Ballesteros, Subramanian, Kawakami, and
  Dyer}]{lample2016neural}
Guillaume Lample, Miguel Ballesteros, Sandeep Subramanian, Kazuya Kawakami, and
  Chris Dyer. 2016.
\newblock \href {https://www.aclweb.org/anthology/N16-1030} {Neural
  architectures for named entity recognition}.
\newblock In \emph{Proceedings of NAACL-HLT}, pages 260--270.

\bibitem[{Ma and Hovy(2016)}]{ma2016end}
Xuezhe Ma and Eduard Hovy. 2016.
\newblock \href {http://dx.doi.org/10.18653/v1/P16-1101} {End-to-end sequence
  labeling via bi-directional {LSTM-CNNs-CRF}}.
\newblock In \emph{Proceedings of the 54th Annual Meeting of the Association
  for Computational Linguistics (Volume 1: Long Papers)}, volume~1, pages
  1064--1074.

\bibitem[{Moreno et~al.(2015)Moreno, Teh, and
  Perez-Cruz}]{moreno_bayesian_2015}
Pablo~G. Moreno, Yee~Whye Teh, and Fernando Perez-Cruz. 2015.
\newblock \href {http://www.jmlr.org/papers/volume16/moreno15a/moreno15a.pdf}
  {{Bayesian} nonparametric crowdsourcing}.
\newblock \emph{Journal of Machine Learning Research}, 16:1607--1627.

\bibitem[{Nguyen et~al.(2017)Nguyen, Wallace, Li, Nenkova, and
  Lease}]{nguyen2017aggregating}
An~T Nguyen, Byron~C Wallace, Junyi~Jessy Li, Ani Nenkova, and Matthew Lease.
  2017.
\newblock \href {http://dx.doi.org/10.18653/v1/P17-1028} {Aggregating and
  predicting sequence labels from crowd annotations}.
\newblock In \emph{Proceedings of the conference. Association for Computational
  Linguistics. Meeting}, volume 2017, page 299. NIH Public Access.

\bibitem[{Paun et~al.(2018)Paun, Carpenter, Chamberlain, Hovy, Kruschwitz, and
  Poesio}]{paun2018comparing}
Silviu Paun, Bob Carpenter, Jon Chamberlain, Dirk Hovy, Udo Kruschwitz, and
  Massimo Poesio. 2018.
\newblock \href {https://www.transacl.org/ojs/index.php/tacl/article/view/1430}
  {Comparing {Bayesian} models of annotation}.
\newblock \emph{Transactions of the Association for Computational Linguistics},
  6:571--585.

\bibitem[{Raykar et~al.(2010)Raykar, Yu, Zhao, Valadez, Florin, Bogoni, and
  Moy}]{Raykar2010}
Vikas.~C. Raykar, Shipeng Yu, Linda~H. Zhao, Gerardo~Hermosillo. Valadez,
  Charles Florin, Luca Bogoni, and Linda Moy. 2010.
\newblock \href {http://www.jmlr.org/papers/v11/raykar10a.html} {Learning from
  crowds}.
\newblock \emph{Journal of Machine Learning Research}, 11:1297--1322.

\bibitem[{Rodrigues et~al.(2013)Rodrigues, Pereira, and
  Ribeiro}]{rodrigues2013learning}
Filipe Rodrigues, Francisco Pereira, and Bernardete Ribeiro. 2013.
\newblock \href {https://core.ac.uk/download/pdf/43577079.pdf} {Learning from
  multiple annotators: distinguishing good from random labelers}.
\newblock \emph{Pattern Recognition Letters}, 34(12):1428--1436.

\bibitem[{Rodrigues et~al.(2014)Rodrigues, Pereira, and
  Ribeiro}]{rodrigues2014sequence}
Filipe Rodrigues, Francisco Pereira, and Bernardete Ribeiro. 2014.
\newblock \href {https://link.springer.com/article/10.1007/s10994-013-5411-2}
  {Sequence labeling with multiple annotators}.
\newblock \emph{Machine learning}, 95(2):165--181.

\bibitem[{Settles(2010)}]{settles2010active}
Burr Settles. 2010.
\newblock \href {https://minds.wisconsin.edu/handle/1793/60660} {Active
  learning literature survey}.
\newblock \emph{Computer Sciences Technical Report 1648, University of
  Wisconsin-Madison}, 52(55-66):11.

\bibitem[{Settles and Craven(2008)}]{settles2008analysis}
Burr Settles and Mark Craven. 2008.
\newblock \href {https://www.cs.cmu.edu/~bsettles/pub/settles.emnlp08.pdf} {An
  analysis of active learning strategies for sequence labeling tasks}.
\newblock In \emph{Proceedings of the conference on empirical methods in
  natural language processing}, pages 1070--1079. Association for Computational
  Linguistics.

\bibitem[{Sheshadri and Lease(2013)}]{sheshadri2013square}
Aashish Sheshadri and Matthew Lease. 2013.
\newblock \href
  {https://www.aaai.org/ocs/index.php/HCOMP/HCOMP13/paper/view/7550} {{SQUARE}:
  A benchmark for research on computing crowd consensus}.
\newblock In \emph{First AAAI Conference on Human Computation and
  Crowdsourcing}.

\bibitem[{Simpson et~al.(2013)Simpson, Roberts, Psorakis, and
  Smith}]{simpsonlong}
Edwin Simpson, Stephen~J. Roberts, Ioannis Psorakis, and Arfon Smith. 2013.
\newblock \href
  {http://www.orchid.ac.uk/eprints/32/1/galaxyZooSN_simpson_etal.pdf} {Dynamic
  {Bayesian} combination of multiple imperfect classifiers}.
\newblock \emph{Intelligent Systems Reference Library series}, Decision Making
  with Imperfect Decision Makers:1--35.

\bibitem[{Srivastava et~al.(2014)Srivastava, Hinton, Krizhevsky, Sutskever, and
  Salakhutdinov}]{srivastava2014dropout}
Nitish Srivastava, Geoffrey Hinton, Alex Krizhevsky, Ilya Sutskever, and Ruslan
  Salakhutdinov. 2014.
\newblock \href {http://jmlr.org/papers/v15/srivastava14a.html} {Dropout: a
  simple way to prevent neural networks from overfitting}.
\newblock \emph{The Journal of Machine Learning Research}, 15(1):1929--1958.

\bibitem[{Tjong Kim~Sang and De~Meulder(2003)}]{tjong2003introduction}
Erik~F Tjong Kim~Sang and Fien De~Meulder. 2003.
\newblock \href {https://www.aclweb.org/anthology/W03-0419} {Introduction to
  the {CoNLL}-2003 shared task: Language-independent named entity recognition}.
\newblock In \emph{Proceedings of the seventh conference on Natural language
  learning at HLT-NAACL 2003-Volume 4}, pages 142--147. Association for
  Computational Linguistics.

\bibitem[{Trautmann et~al.(2019)Trautmann, Daxenberger, Stab, Sch{\"u}tze, and
  Gurevych}]{trautmann2019robust}
Dietrich Trautmann, Johannes Daxenberger, Christian Stab, Hinrich Sch{\"u}tze,
  and Iryna Gurevych. 2019.
\newblock \href {https://arxiv.org/abs/1904.09688} {Robust argument unit
  recognition and classification}.
\newblock \emph{arXiv preprint arXiv:1904.09688}.

\bibitem[{Venanzi et~al.(2014)Venanzi, Guiver, Kazai, Kohli, and
  Shokouhi}]{venanzi2014community}
Matteo Venanzi, John Guiver, Gabriella Kazai, Pushmeet Kohli, and Milad
  Shokouhi. 2014.
\newblock \href {https://doi.org/10.1145/2566486.2567989} {Community-based
  {Bayesian} aggregation models for crowdsourcing}.
\newblock In \emph{23rd international conference on World wide web}, pages
  155--164.

\bibitem[{Venanzi et~al.(2016)Venanzi, Guiver, Kohli, and
  Jennings}]{venanzi2016time}
Matteo Venanzi, John Guiver, Pushmeet Kohli, and Nicholas~R Jennings. 2016.
\newblock \href {https://doi.org/10.1613/jair.5175} {Time-sensitive {Bayesian}
  information aggregation for crowdsourcing systems}.
\newblock \emph{Journal of Artificial Intelligence Research}, 56:517--545.

\bibitem[{Viterbi(1967)}]{viterbi1967error}
Andrew Viterbi. 1967.
\newblock \href {https://ieeexplore.ieee.org/abstract/document/1054010} {Error
  bounds for convolutional codes and an asymptotically optimum decoding
  algorithm}.
\newblock \emph{IEEE transactions on Information Theory}, 13(2):260--269.

\bibitem[{Whitehill et~al.(2009)Whitehill, Wu, Bergsma, Movellan, and
  Ruvolo}]{whitehill2009whose}
Jacob Whitehill, Ting-fan Wu, Jacob Bergsma, Javier~R Movellan, and Paul~L
  Ruvolo. 2009.
\newblock \href
  {https://papers.nips.cc/paper/3644-whose-vote-should-count-more-optimal-integration-of-labels-from-labelers-of-unknown-expertise}
  {Whose vote should count more: Optimal integration of labels from labelers of
  unknown expertise}.
\newblock In \emph{Advances in neural information processing systems}, pages
  2035--2043.

\bibitem[{Xiong et~al.(2011)Xiong, Barash, and Frey}]{xiong2011bayesian}
Hui~Yuan Xiong, Yoseph Barash, and Brendan~J Frey. 2011.
\newblock \href {https://doi.org/10.1093/bioinformatics/btr444} {Bayesian
  prediction of tissue-regulated splicing using rna sequence and cellular
  context}.
\newblock \emph{Bioinformatics}, 27(18):2554--2562.

\bibitem[{Zhang(2004)}]{zhang2004optimality}
Harry Zhang. 2004.
\newblock \href {https://www.aaai.org/Papers/FLAIRS/2004/Flairs04-097.pdf} {The
  optimality of na\"ive {Bayes}}.
\newblock In \emph{Proceedings of the Seventeenth International Florida
  Artificial Intelligence Research Society Conference, FLAIRS 2004}. AAAI
  Press.

\end{thebibliography}

\appendix
\section{Discussion of Variational Approximation}

In Equation 12,
each subset of latent variables has a variational factor of the form 
$\ln q(z) = \mathbb{E}[\ln p(z | \bs c, \neg z)]$, 
where $\neg z$ contains all the other latent variables excluding $z$.
We perform approximate inference by
%we optimize Equation \ref{eq:vb_posterior} 
using coordinate ascent to update each variational factor, $q(z)$, in turn,
taking expectations with respect to the current estimates of the other variational factors.
%(see Algorithm \ref{al:vb_bac}).
Each iteration reduces the KL-divergence between the true and approximate posteriors
of Equation 12, and hence optimises a lower bound on the 
log marginal likelihood, also called the evidence lower bound or ELBO
~\cite{Bishop2006,attias_advances_2000}.

\paragraph{Conjugate distributions: }The prior distributions chosen for our generative model are conjugate to the distributions over the
latent variables and model parameters, 
meaning that each $q(z)$ is the same type of distribution
as the corresponding  prior distribution defined in Section 4.
The parameters of each variational distribution are computed in terms  of 
expectations over the other subsets of variables.

\begin{figure*}[t]
\begin{minipage}[b][1cm][l]{0.2\textwidth} 
BSC-seq, \\
previous label = I:
\end{minipage}
  \includegraphics[width=0.8\textwidth, clip=True, trim=10 0 0 10]{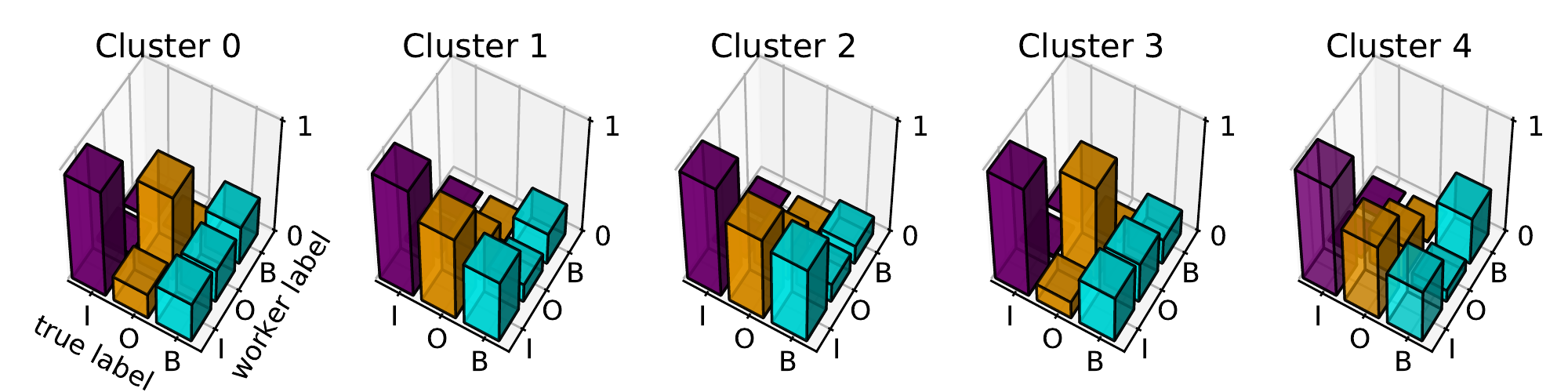}
\\
\begin{minipage}[b][1cm][l]{0.2\textwidth} 
BSC-seq, \\
previous label = O:
\end{minipage}
  \includegraphics[width=0.8\textwidth, clip=True, trim=10 0 0 0]{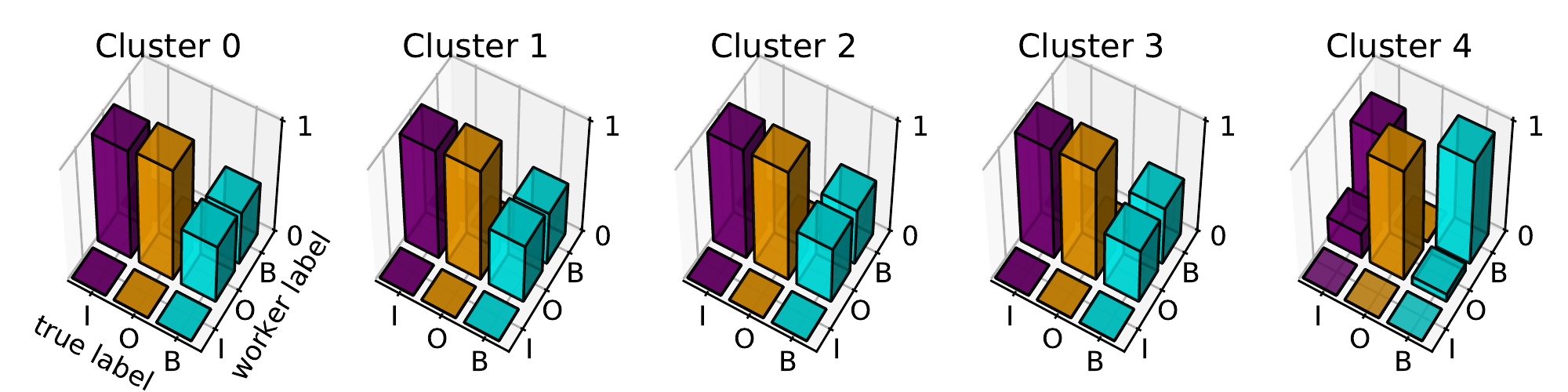}
\\
\begin{minipage}[b][1cm][l]{0.2\textwidth} 
BSC-seq,\\
 previous label = B:
\end{minipage}
  \includegraphics[width=0.8\textwidth, clip=True, trim=10 0 0 0]{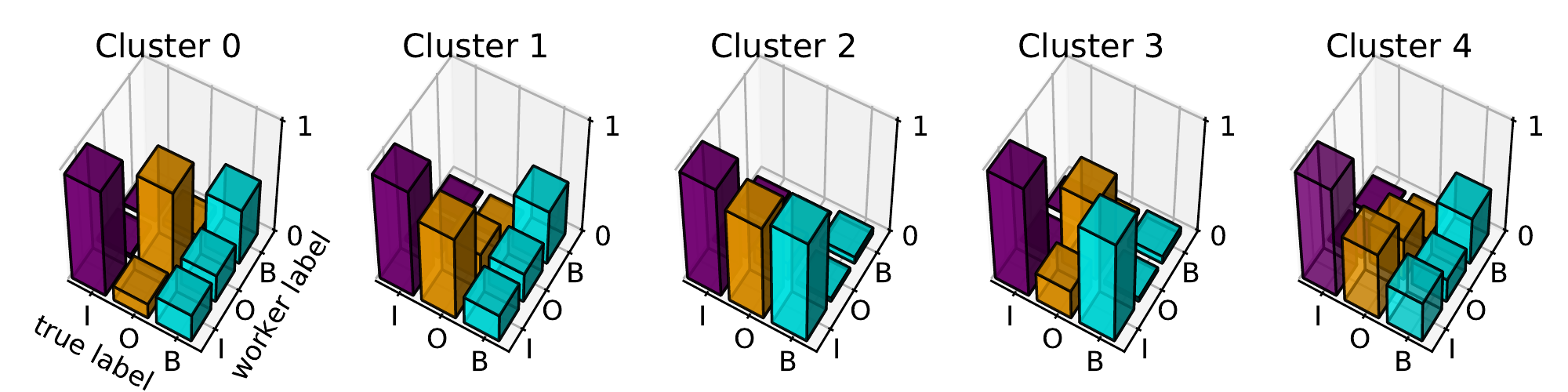}
\\
\caption{Clusters of annotators in PICO represented by the 
mean confusion matrices from BSC-seq. Heights of bars indicate likelihood of a worker (or annotator) label given the true label.
}
\label{fig:anno_models_2}
\end{figure*}

\section{Update Equations for the Forward-Backward Algorithm}

For the true labels, $\bs t$, the variational factor is:
 \begin{flalign}
& \ln q(\bs t_n) \!=\! 
% \sum_{n=1}^N \sum_{\tau=1}^{L_n} 
% \sum_{s=1}^S \mathbb{E}%_{\bs B,\bs d^{(s)}} \!\left[ 
% \ln \!B^{(s)}\!\!\left(t_{n,\tau},d_{n,\tau}^{(s)},d_{n,\tau\!-\!1}^{(s)}\!\right) %\right]
% %\bigg\{ \mathbb{E}%_{\bs T} \left[ 
% %\ln T_{t_{n,\tau\!-\!1}, t_{n,\tau}} %\right] 
% &&\nonumber \\
%+ 
\sum_{n=1}^N \sum_{\tau=1}^{L_n} \sum_{k=1}^K  \mathbb{E}%_{\bs A} \left[ 
\ln \!A^{(k)}\left(t_{n,\tau},c_{n,\tau}^{(k)},c_{n,\tau-1}^{(k)}\right) %\right]  
&\nonumber\\
&+ \mathbb{E}\ln T_{t_{n,\tau-1}, t_{n,\tau}}+\mathrm{const}. & \label{eq:qstar_t}
 \end{flalign}
 
% TODO move all this to the appendix
 The forward-backward algorithm consists of two passes. 
 The \emph{forward pass} for each document, $n$, starts from $\tau=1$
 and computes:% for each value of $\tau$:
 % the posterior given crowdsourced annotations for tokens $\leq\tau$. 
 \begin{flalign}
   & \ln r^{-}_{n,\tau,j} = \ln \sum_{\iota=1}^J \left\{ r^{-}_{n,\tau-1,\iota} \mathrm{e}^{\mathbb{E}\ln T_{\iota,j}} \right\} + ll_{n,\tau}(j), & \nonumber \\
% \end{flalign}
%  where the log likelihood $ll(j,n,\tau)$ of the annotations for token $\tau$ in document $n$ giventr
%  label $j$ is:
% \begin{flalign} 
   & ll_{n,\tau}(j) = \sum_{k=1}^K \mathbb{E}%_{\bs A}
   \ln A^{(k)}\left(j, c_{n,\tau}^{(k)}, c_{n,\tau\!-\!1}^{(k)} \right) & 
   %+  \sum_{s=1}^S
%    & \nonumber \\
%    &  \sum_{i=1}^J\sum_{\iota=1}^J \mathbb{E}%_{\bs B}
%    \ln B^{(s)} \!\left(j, i, \iota \right)  
%    \hat{d}_{n,\tau}^{(s)}(i) \hat{d}_{n,\tau-1}^{(s)}(\iota), & 
 \end{flalign}
 %where $\hat{d}_{n,\tau}^{(s)}(i)$ is %the probability of label $d_{n,\tau}^{(s)}$ produced 
% by the sequence tagger $s$, which we explain in more detail below (see Equation \ref{eq:hatp}).
%defined below in Equation \ref{eq:hatp}, and 
where $r^{-}_{n,0,\iota}  = 1$ where $\iota=$`O' and $0$ otherwise.
The \emph{backwards pass} starts from $\tau=L_n$ and scrolls backwards, computing:
%at each token computing the likelihoods of the annotations from $\tau+1$ to $L_n$:
 \begin{flalign}
  & \ln \lambda_{n,L_n,j} = 0, \hspace{1cm}
   \ln \lambda_{n,\tau,j} = \ln\sum_{\iota=1}^J \exp \big\{ 
   & \nonumber \\
& \ln \lambda_{i,\tau+1,\iota} + \mathbb{E}\ln T_{j,\iota} + ll_{n,\tau+1}(\iota) \big\} .&
 \end{flalign}
% To avoid $r^{-}_{n,\tau,j}$ and $\lambda_{n,\tau,j}$ becoming too small over a long sequence, we normalize them after each iteration of the forward and backward pass
% by dividing by their sum over $j$.
 By %taking the exponents and 
 applying Bayes' rule, we arrive at $r_{n,\tau,j}$ and $s_{n,\tau,j,\iota}$:
 \begin{flalign}
  & r_{n,\tau,j} = \frac{r^{-}_{n,\tau,j}\lambda_{n,\tau,j}}{\sum_{j'=1}^J r^{-}_{n,\tau,j'}\lambda_{n,\tau,j'}} &\\
%} {\sum_{\iota=1}^J \sum_{\iota'=1}^J  
%  r^{-}_{n,\tau\!-\!1,\iota}\lambda_{n,\tau,\iota'} \exp\mathbb{E}[\ln T_{\iota,\iota'}] 
% + ll(\iota',n,\tau)  } . &
& s_{n,\tau,j,\iota} = \frac{ \tilde{s}_{n,\tau,j,\iota} }{ \sum_{j'=1}^J\sum_{\iota'=1}^J  \tilde{s}_{n,\tau,j',\iota'} } & \\
  & \tilde{s}_{n,\tau,j,\iota} =  r^{-}_{n,\tau-1,j} \lambda_{n,\tau,\iota} \exp\{\mathbb{E}\ln T_{j,\iota}
+ ll_{n,\tau}(\iota)\}. & \nonumber 
 \end{flalign}
 
 Each row of the transition matrix has the factor:
\begin{flalign}
& \ln q(\bs T_{j}) 
  %= \sum_{\iota=1}^J N_{j,\iota}  + \ln \mathrm{Dir}(\bs T_j | \bs\gamma_j) + \mathrm{const} & \nonumber\\
= \ln \mathrm{Dir}\left(\left[ N_{j,\iota} + \gamma_{j,\iota}, \forall \iota \in \{1,..,J\} \right]\right), &
\end{flalign}
where $N_{j,\iota} = \sum_{n=1}^N \sum_{\tau=1}^{L_n}  s_{n,\tau,j,\iota}$ is the expected number of times that label $\iota$ follows label $j$.  
%The variational factor $q(\bs t)$ requires the following expectation:
The forward-backward algorithm requires expectations of $\ln \bs T$ that can be computed using standard equations for a Dirichlet distribution:
 \begin{flalign}
& \mathbb{E}\ln T_{j,\iota} = \Psi\!\left(N_{j,\iota} \!\!+ \gamma_{j,\iota}\right) 
 - \Psi\!\left(\sum_{\iota=1}^J (N_{j,\iota} \!\!+ \gamma_{j,\iota}) \!\right), &
\end{flalign}
 where $\Psi$ is the digamma function.
 
%\textbf{Variational factors for} $\bs A$ and $\bs B$:
The variational factor for each annotator model is a distribution over its parameters, 
which differs between models.
For \emph{seq}, the variational factor is:
 \begin{flalign}
  & \ln q\!\left(\! A^{(k)}\!\right) %= \sum_{j=1}^J  \sum_{l=1}^J \bigg\{ \sum_{m=1}^J N_{j,l,m}^{(k)}\ln\pi_{j,l,m}^{(k)} & \nonumber\\
 % & \hspace{2.7cm} 
 % + \ln p\left(\bs\pi_{j,l}^{(k)} | \bs \alpha_{j,l}^{(k)} \right) \bigg\} + \mathrm{const}, & \nonumber \\
  \!=\! \sum_{j=1}^J \! \sum_{l=1}^J \!\mathrm{Dir}\! \left(\left[ \bs N_{j,l,m}^{(k)} \! 
 %+ \alpha_{j,l,m}^{(k)}, \! 
 %\right.\right.\nonumber&\\
%&  \hspace{3.5cm} \left.\left.
\forall m \! \in \! \{1,..,J\} \!\right] \right) & \nonumber \\
& N^{(k)}_{j,l,m} \!\!=\!  \alpha_{j,l,m}^{(k)} \!\!\! + \!\sum_{n=1}^N \!\sum_{\tau=1}^{L_n} \!
r_{n,\tau,j} \delta_{l,c^{(k)}_{n,\tau\!-\!1}}\!\delta_{m,c^{(k)}_{n, \!\tau}}, \!& 
\end{flalign}
 where $\delta$ is the Kronecker delta. 
% For the \emph{CM} model, the variational factor is simplified to:
%  \begin{flalign}
%   & \ln q\left( A^{(k)}\right) = \sum_{j=1}^J  \mathrm{Dir} \bigg( \bigg[ \sum_{n=1}^N \sum_{\tau=1}^{L_n} r_{n,\tau,j} \delta_{m,c^{(k)}_{n,\tau}} 
%   & \nonumber \\ 
% & \hspace{2.0cm} + \alpha_{j,m}^{(k)}, \! \forall m \! \in \! \{1,..,J\} \bigg] \bigg) .
% \end{flalign}
For \emph{CM}, \emph{MACE}, \emph{CV} and \emph{acc}, the factors follow a similar pattern of summing pseudo-counts of correct and incorrect answers. 
%For reasons of space, we omit the equations for these variants. 

\section{Visualising Annotator Models}

Figure \ref{fig:anno_models_2} provides an alternative visualisation of the \textit{seq} models inferred by BSC-seq for annotators in the PICO dataset.
The annotators were clustered as described in Section 6 of the main paper,
and the mean confusion matrices for each cluster are plotted in Figure \ref{fig:anno_models_2} using 3D plots to emphasise the differences between 
the likelihoods of annotators in each cluster providing a particular label 
given the true label value.

\end{document}